\documentclass[11pt]{article}

\usepackage{epsfig,amsmath,latexsym,amssymb}
\usepackage{graphicx}
\usepackage{lscape}
\usepackage{picture, eso-pic, tikz} %% for gray boxes
\usepackage{dsfont}
\usepackage{listings}
\usepackage{changes}
\usepackage{hyperref}
\usepackage{bbding}
\usepackage{url}
\usepackage{booktabs}

\renewcommand{\baselinestretch}{1.1}
\def\R{{\mathbb R}}  %%
\def\p{{\mathbb P}}  %% 
\def\E{{\mathbb E}}  %
\def\Beweis{\footnotesize}

\newcommand{\Remm}[1]{}
\newtheorem{theo}{Theorem}[section]
\newtheorem{lemma}[theo]{Lemma}

\newtheorem{model ass}[theo]{Model Assumptions}
\newtheorem{ass}[theo]{Assumptions}

\newtheorem{example}[theo]{Example}

\newtheorem{rem}[theo]{Remark}
\newtheorem{rems}[theo]{Remarks}

\def\EndProof{\hfill {\scriptsize $\Box$}}
\def\EndExample{\hfill {\scriptsize $\blacksquare$}}

\numberwithin{equation}{section}

\definecolor{MyGray}{rgb}{0.92,0.92,0.92}


\definecolor{British racing}{rgb}{0.0, 0.5, 0.0}

\def\bX{\boldsymbol{X}}
\def\b0{\boldsymbol{0}}

\def\b0{\boldsymbol{0}}

\usepackage{todonotes}
\newcommand{\Comments}{1}
\newcommand{\mynote}[2]{\ifnum\Comments=1\textcolor{#1}{#2}\fi}
\newcommand{\mytodo}[2]{\ifnum\Comments=1%
  \todo[linecolor=#1!80!black,backgroundcolor=#1,bordercolor=#1!80!black]{#2}\fi}

\ifnum\Comments=1               % fix margins for todonotes
\fi

\ifnum\Comments=1               % fix margins for todonotes
\fi

\ifnum\Comments=1               % fix margins for todonotes
\fi

\begin{document}
\author{Alexej Brauer\footnote{Actuarial Department, Allianz Versicherungs-AG, Munich, brauer.alexej@gmail.com} \and
Mario V.~W\"uthrich\footnote{Department of Mathematics, ETH Zurich,
mario.wuethrich@math.ethz.ch}}

\date{Version of \today}
%\date{}
\title{Gini Score under Ties and Case Weights}
\maketitle

\begin{abstract}
\noindent  
The Gini score is a popular tool in statistical modeling and machine learning for model validation and model selection. It is a purely rank based score that allows one to assess risk rankings. The Gini score for statistical modeling has mainly been used in a binary context, in which it has many equivalent reformulations such as the receiver operating characteristic (ROC) or the area under the curve (AUC). In the actuarial literature, this rank based score for binary responses has been extended to general real-valued random variables using Lorenz curves and concentration curves. While these initial concepts assume that the risk ranking is generated by a continuous distribution function, we discuss in this paper how the Gini score can be used in the case of ties in the risk ranking. Moreover, we adapt the Gini score to the common actuarial situation of having case weights.
\medskip


\end{abstract}

\section{Introduction}
The Gini score is a popular statistical tool in model validation.
The Gini score has originally been introduced and used for binary responses $Y \in \{0,1\}$, and there are many equivalent formulations of the (binary) Gini score such as the receiver operating curve (ROC) and the area under the curve (AUC); see, e.g., \cite{Bamber}, \cite{Hanley} and \cite{Fawcett}. These different formulations are also equivalent to the Wilcoxon--Mann-Whitney's $U$ statistic, see \cite{Hanley}, \cite{DeLong}, \cite{Byrne}, and to \cite{Somers}'s $D$, see \cite{Newson}. Thus, there are at least five equivalent formulations of the Gini score in a binary context, and there is a broad literature on its behavior which is well understood.

When it comes to general real-valued responses, things become more difficult, and definitions and results on the Gini score are mainly found in the credit risk and actuarial literature. In this stream of literature, the Gini score has been introduced by \cite{GourierouxJasiak}, \cite{Frees1, Frees2}. Furthermore, in the real-valued setting the Gini score is studied in much detail in \cite{DenuitSznajderTrufin} and \cite{DenuitTrufin}.

The Gini score is a statistic that assesses whether a given {\it risk ranking} is correct. Assume we have a sample $(Y_i,\mu_i, \widehat{\mu}_i)_{i=1}^n$ with true means $\E[Y_i]=\mu_i$ and corresponding estimates $\widehat{\mu}_i$, for $1\le i \le n$. The property assessed by the Gini score is whether the true means $(\mu_i)_{i=1}^n$ and their estimates $(\widehat{\mu}_i)_{i=1}^n$ provide the same risk ranking, the maximal Gini score being attained if both risk rankings are identical. Thus, the Gini score is a purely rank based concordance statistic, and it does not assess whether the levels of the estimates $(\widehat{\mu}_i)_{i=1}^n$ are correct, in particular, the shifted estimates $(a+\widehat{\mu}_i)_{i=1}^n$, for $a \in \R$, provide the identical risk ranking and, thus, they have the same Gini score. Even if we do not have the correct levels (also called calibration), assessing the risk ranking is often of central interest, e.g., the correct risk ranking is a crucial assumption in the application of an isotonic regression, see \cite{ZiegelW}, and the universal inference calibration test of \cite{DelongW} is based on the assumption of having the correct risk ranking.

The Gini score, as it was introduced in \cite{DenuitSznajderTrufin}, is based on the (standard) assumption that the risk ranking follows a continuous random variable which, in particular, implies that there are no ties in the risk ranking $(\widehat{\mu}_i)_{i=1}^n$. However, in numerous practical applications, there are ties in the estimates $(\widehat{\mu}_i)_{i=1}^n$, e.g., if these estimates have been generated by gradient boosting machines (GBMs), then by the very nature of their construction through regression trees, there are naturally ties in the resulting risk rankings. Similarly, in large insurance portfolios, it is likely that there are multiple insurance policyholders that have the same covariates and, thus, receive the same prediction for their claims (being based on these covariates). In such cases, there result ties in the samples, and the (classical) Gini score needs to be adapted to deal with these ties.

Our two main contributions are the following.  (1) First, we discuss the treatment of ties in the risk scoring. Handling ties turns out to be surprisingly sensitive, and its treatment is crucial in a proper model selection process. (2) Second, we discuss how to deal with case weights. Often responses are supported be varying case weights, e.g., claims frequencies come together with time exposures. We discuss how such exposures and case weights can be integrated into Gini scores in the model selection process.

\medskip

{\bf Organization.} The Gini score is built on two different curves, these are the Lorenz curve that builds the nominator of the Gini score and the concentration curve (cumulative accuracy profile) that builds the numerator of the Gini score. We discuss the Lorenz curve and the treatment of ties in the risk ranking in Section \ref{sec: Lorenz curve}, and correspondingly for the concentration curve in Section \ref{Cumulative accuracy profile}. Section \ref{case weights section} discusses the situation under case weights, and in Section \ref{sec: Real data example} we present a real data example that is based on claims frequencies which naturally include exposures (case weights). Finally, in Section \ref{sec: Conclusion} we conclude.

\section{Lorenz curve}
\label{sec: Lorenz curve}
The Gini score is motivated by an economic concept introduced by
\cite{Gini0, Gini}. Based on the \cite{Lorenz} curve, the Gini score measures the disparity of the wealth distribution within a given population. The bigger the Gini score, the bigger the inequality in wealth distribution. This concept has been adapted to predictive modeling, by measuring how well a given regression function discriminates the responses. 
In this section, we are going to introduce the \cite{Lorenz} curve, before diving into Gini scoring in the next section.

\subsection{Mirrored Lorenz curve}
Select a non-negative random variable $Y \sim F_Y$ with positive finite first moment $\E[Y]\in (0,\infty)$. The Lorenz curve is given by the function, see \cite[Definition 3.2]{DenuitSznajderTrufin},
\begin{equation}\label{Lorenz A1}
\alpha \in (0,1)~ \mapsto ~ \frac{1}{\E[Y]} \,\E \left[ Y \,\mathds{1}_{\{Y \le F_Y^{-1}(\alpha)\}}\right],
\end{equation}
where $F_Y^{-1}$ is the left-continuous generalized inverse of the distribution $F_Y$, given by
\begin{equation*}
F_Y^{-1}(\alpha) = \inf \left\{ y \in \R;\, F_Y(y)\ge \alpha \right\}.
\end{equation*}
For the machine learning version of the Gini score, we use the mirrored version of the Lorenz curve (the upper tail), given by 
\begin{equation}\label{Lorenz A2}
\alpha \in (0,1)~ \mapsto ~ L_Y(\alpha) = \frac{1}{\E[Y]} \,\E \left[ Y \,\mathds{1}_{\{Y > F_Y^{-1}(1-\alpha)\}}\right]
=
1- \frac{1}{\E[Y]} \,\E \left[ Y \,\mathds{1}_{\{Y \le F_Y^{-1}(1-\alpha)\}}\right].
\end{equation}
By a slight abuse of terminology, we call this mirrored version
the {\it Lorenz curve} in this manuscript. Moreover, we calibrate the Lorenz curve $L_Y(\cdot)$ at the boundary of the open interval $(0,1)$ to 
$L_Y(0)=0$ and $L_Y(1)=1$.

\begin{example}[continuous example]\normalfont
\label{LN exact}
We consider a continuous distribution example. Assume the random variable $Y$ has a log-normal distribution
$\log(Y) \sim {\cal N}(\mu, \sigma^2)$. The inverse is given
by
\begin{equation*}
F_Y^{-1}(1-\alpha) =  \exp \left(\mu + \sigma \Phi^{-1}(1-\alpha)\right),
\end{equation*}
where $\Phi$ is the standard Gaussian distribution. This gives for the Lorenz curve
\begin{equation*}
\alpha \in (0,1)~ \mapsto ~ L_Y(\alpha) = 1- \Phi \left(\frac{\log \left(F_Y^{-1}(1-\alpha)\right) - (\mu+\sigma^2)}{\sigma}\right)
= 1- \Phi \left( \Phi^{-1}(1-\alpha) - \sigma\right).
\end{equation*}

\begin{figure}[htb!]
\begin{center}
\begin{minipage}[t]{0.45\textwidth}
\begin{center}
\includegraphics[width=\textwidth]{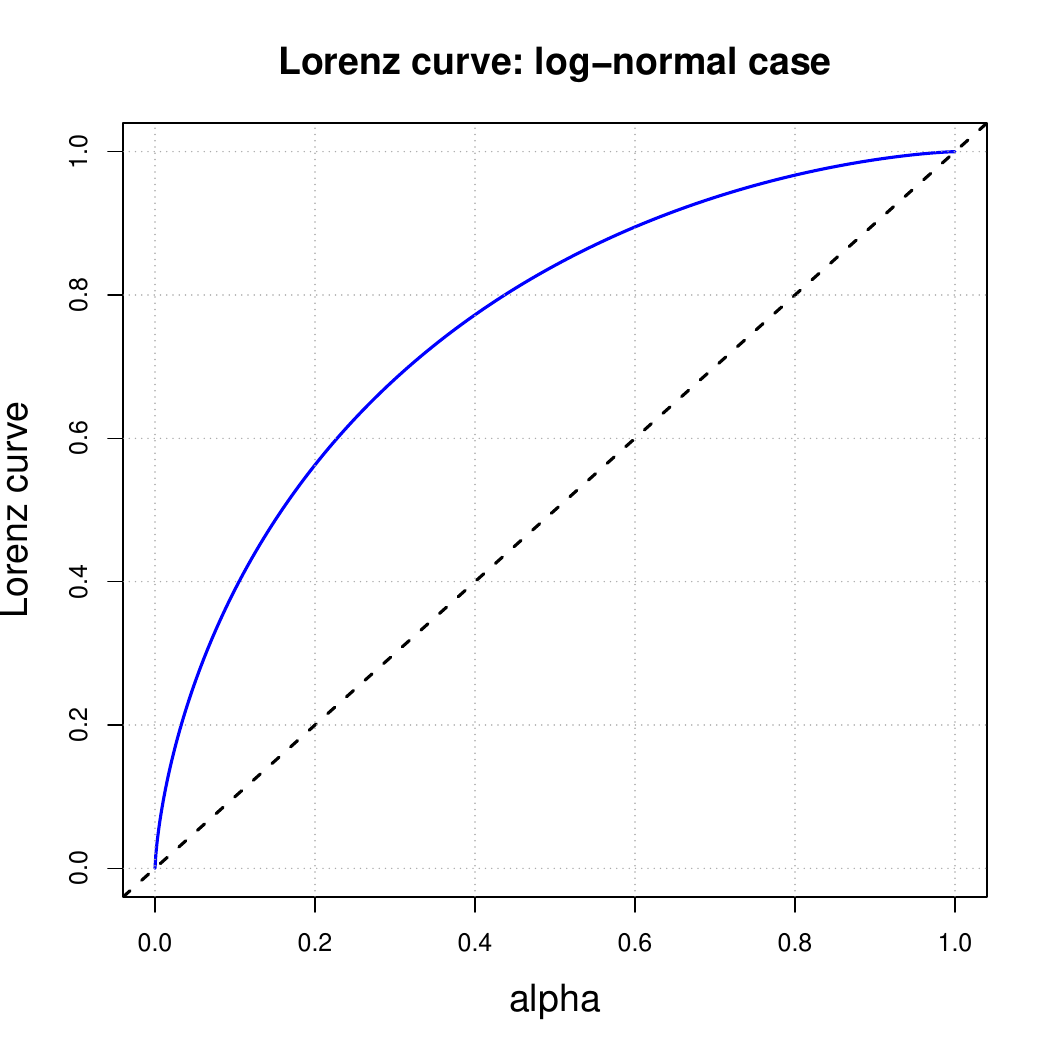}
\end{center}
\end{minipage}
\begin{minipage}[t]{0.45\textwidth}
\begin{center}
\includegraphics[width=\textwidth]{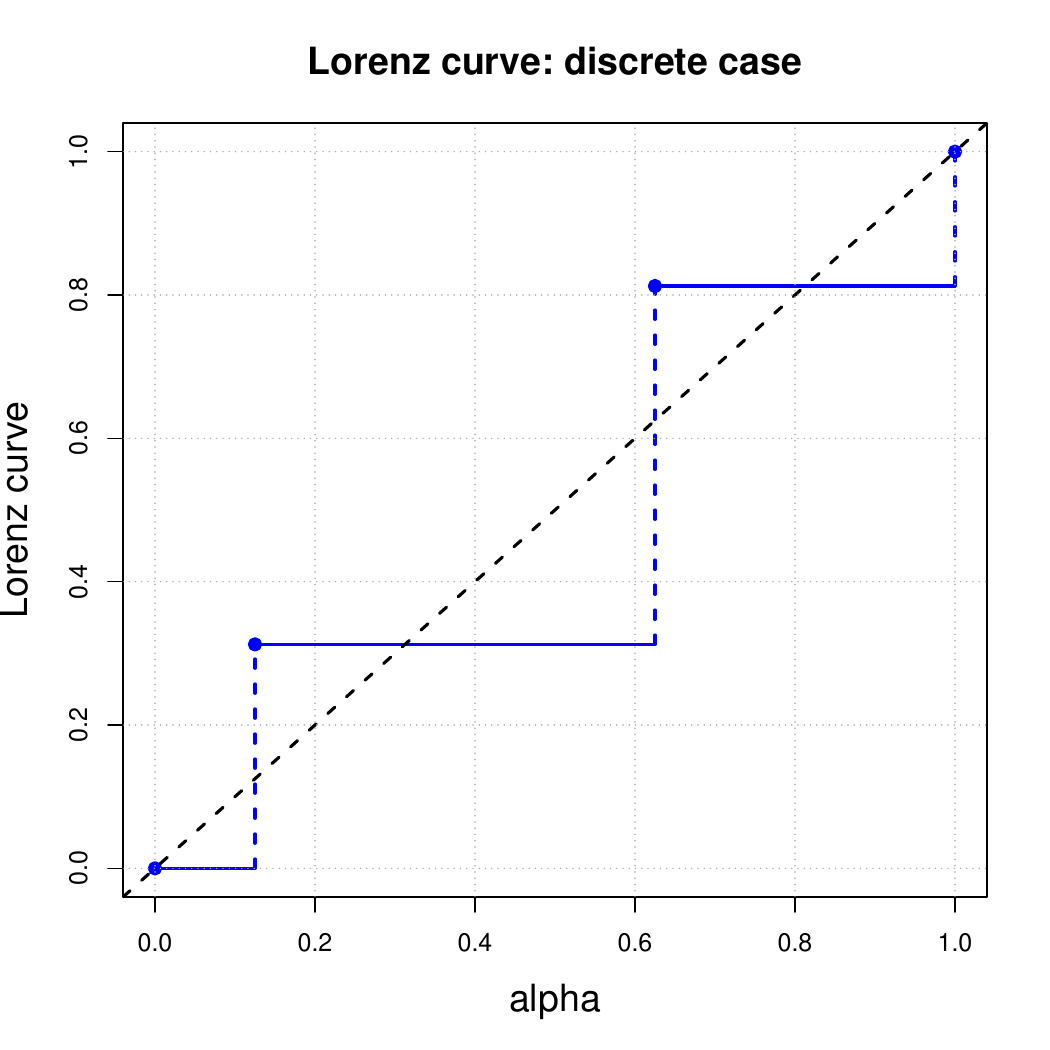}
\end{center}
\end{minipage}
\end{center}
\caption{ Lorenz curve: (lhs) log-normal case with $\sigma=1$; (rhs) discrete example.}
\label{Lorenz exact}
\end{figure}
This Lorenz curve of this log-normal example is illustrated in Figure \ref{Lorenz exact} (lhs) for $\sigma=1$.
\EndExample
\end{example}

\begin{example}[discrete example]\normalfont
\label{discrete exact} 
We consider a discrete example. Assume the random variable $Y$ takes the following three values 
\begin{equation}\label{true discrete distribution}
Y = \left\{
\begin{array}{ll}
1/2 & \text{ with probability $p_1=3/8$,}\\
1 & \text{ with probability $p_2=1/2$,}\\
5/2 & \text{ with probability $p_3=1/8$.}
\end{array}
\right.
\end{equation}
This random value has mean $\E[Y]=1$ and its distribution $F_Y$ is a step function with steps in $y_k\in \{1/2,1,5/2\}$ having step sizes $p_k\in \{3/8,1/2,1/8\}$, for $1\le k \le 3$. We compute the left-continuous generalized inverse for $p \in (0,1]$
\begin{equation*}
F_Y^{-1}(p) = \left\{
\begin{array}{ll}
1/2 & \text{ for $p \in (0,3/8]$,}\\
1 & \text{ for $p \in (3/8,7/8]$,}\\
5/2 & \text{ for $p \in (7/8,1]$.}
\end{array}
\right.
\end{equation*}
This gives us the Lorenz curve
\begin{equation*}
\alpha \in [0,1]~ \mapsto ~ L_Y(\alpha)=  \sum_{k=1}^3 p_k \, y_k\, \mathds{1}_{\{y_k > F_Y^{-1}(1-\alpha)\}}
= \left\{
\begin{array}{ll}
0 & \text{ for $\alpha \in [0,1/8)$,}\\
5/16 & \text{ for $\alpha \in [1/8,5/8)$,}\\
13/16 & \text{ for $\alpha \in [5/8,1)$,}\\
1 & \text{ for $\alpha=1$.}
\end{array}
\right.
\end{equation*}
This Lorenz curve is illustrated in Figure \ref{Lorenz exact} (rhs), and we notice that it is a step function in this discrete case.
\EndExample
\end{example}

Examples \ref{LN exact} and \ref{discrete exact} show two exact examples that have been computed from the true distribution $F_Y$ of the response $Y$. In a next step, we consider the situation where the true distribution is unknown, and we are (only) equipped with an i.i.d.~sample $(Y_i)_{i=1}^n$ that has been generated from $F_Y$.
This allows us to define the empirical distribution
\begin{equation*}
\widehat{F}_n(y) = \frac{1}{n}\, \sum_{i=1}^n \mathds{1}_{\{ Y_i \le y \}},
\qquad \text{ for $y \in \R$.}
\end{equation*}
The Glivenko--Cantelli result tells us that the empirical distribution 
$\widehat{F}_n$ converges to the true distribution $F_Y$ uniformly, a.s., that
is $\lim_{n\to \infty}\|\widehat{F}_n-F_Y\|_\infty=0$, a.s. This justifies to take the empirical distribution as an approximation to the true one (for large $n$).
This empirical distribution motivates to estimate the Lorenz curve $L_Y$ by its empirical counterpart
\begin{equation}\label{general empirical Lorenz}
\alpha \in (0,1)~ \mapsto ~
\widehat{L}_n(\alpha) = \frac{1}{\frac{1}{n}\sum_{i=1}^n Y_i}\,
\frac{1}{n}\, \sum_{i=1}^n Y_i \,\mathds{1}_{\{Y_i > \widehat{F}_n^{-1}(1-\alpha)\}},
\end{equation}
again we set at the boundary of the unit interval
$\widehat{L}_n(0)=0$ and $\widehat{L}_n(1)=1$.
This gives us a step function, completely analogously to Example \ref{discrete exact}, with steps sizes determined by the observations $(Y_i)_{i=1}^n$.

In case of {\it no ties} in the observations, we can build the strict order statistics
$Y_{(1)}>\ldots > Y_{(n)}$, and all step heights in the empirical distribution $\widehat{F}_n$
are equal to $1/n$. This then allows one to express the empirical  Lorenz curve by (in the case {\it without ties})
\begin{equation}\label{no ties empirical Lorenz}
\alpha \in (0,1)~ \mapsto ~
\widehat{L}_n(\alpha) = \frac{1}{\frac{1}{n}\sum_{i=1}^n Y_i}\,
\frac{1}{n}\, \sum_{i=1}^{n-\lceil (1-\alpha)n\rceil} Y_{(i)},
\end{equation}
an empty sum is set equal to zero.

\begin{example}[continuous example: empirical version]\normalfont
\label{LN empirical}
We revisit the log-normal case of Example \ref{LN exact}. The log-normal example has a continuous distribution. As a consequence, we cannot have ties in the observed sample 
$(Y_i)_{i=1}^n$. Therefore, formula \eqref{no ties empirical Lorenz} applies.
\begin{figure}[htb!]
\begin{center}
\begin{minipage}[t]{0.45\textwidth}
\begin{center}
\includegraphics[width=\textwidth]{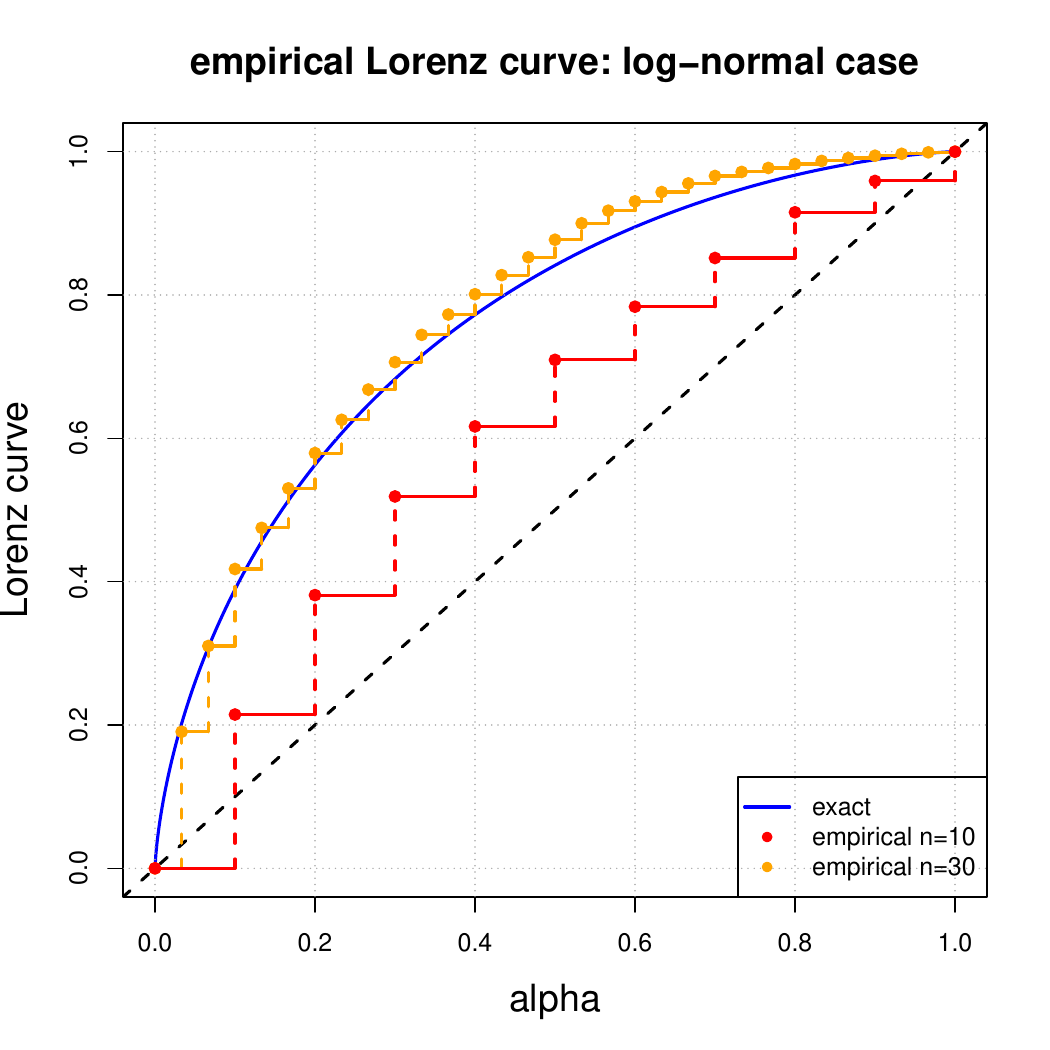}
\end{center}
\end{minipage}
\begin{minipage}[t]{0.45\textwidth}
\begin{center}
\includegraphics[width=\textwidth]{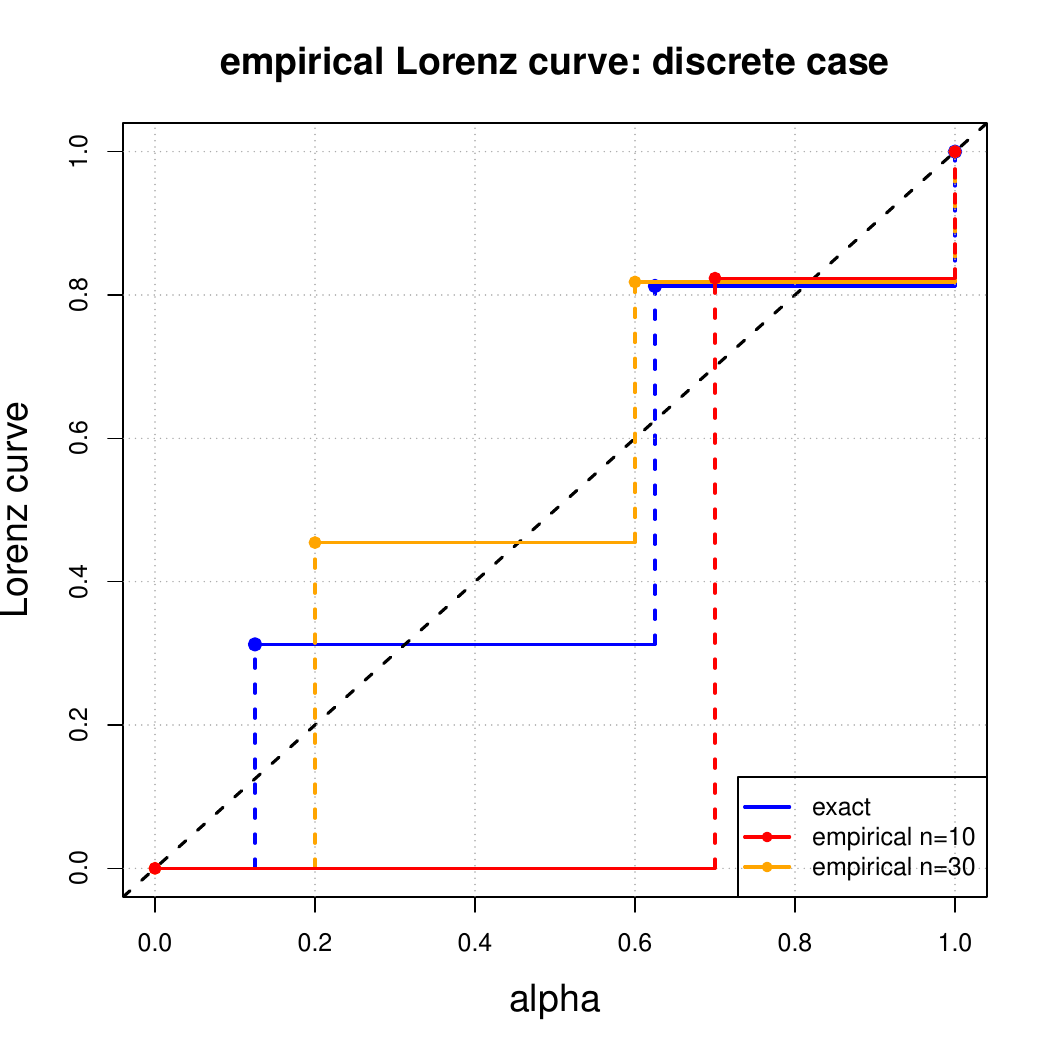}
\end{center}
\end{minipage}
\end{center}
\caption{Empirical Lorenz curve for sample sizes $n=10,30$: (lhs) empirical log-normal case; (rhs) empirical discrete case.}
\label{empirical Lorenz}
\end{figure}

Figure \ref{empirical Lorenz} (lhs) gives two empirical examples of different sample sizes $n=10,30$. Naturally, if we resample the observations $(Y_i)_{i=1}^n$ we get different results (on finite samples), but the results converge to the true Lorenz curve, a.s., as the sample size goes to infinity.
\EndExample
\end{example}

\begin{example}[discrete example: empirical version]\normalfont
\label{discrete empirical}
We revisit the discrete Example \ref{discrete exact}. We simulate two samples of sample sizes $n=10,30$ from the true discrete distribution \eqref{true discrete distribution}. Figure \ref{empirical Lorenz} (rhs) shows the results. The smaller sample only contains the two labels $\{1/2, 1\}$, and we obtain a step function with two steps for the empirical Lorenz curve $\widehat{L}_n$. In the bigger sample all three labels $\{1/2, 1, 5/2\}$ occur, and the empirical Lorenz curve starts to approximate the true one more closely. These empirical Lorenz curves were computed from \eqref{general empirical Lorenz}.
\EndExample
\end{example}

\subsection{Linearly interpolated (empirical) Lorenz curve}
\begin{figure}[htb!]
\begin{center}
\begin{minipage}[t]{0.45\textwidth}
\begin{center}
\includegraphics[width=\textwidth]{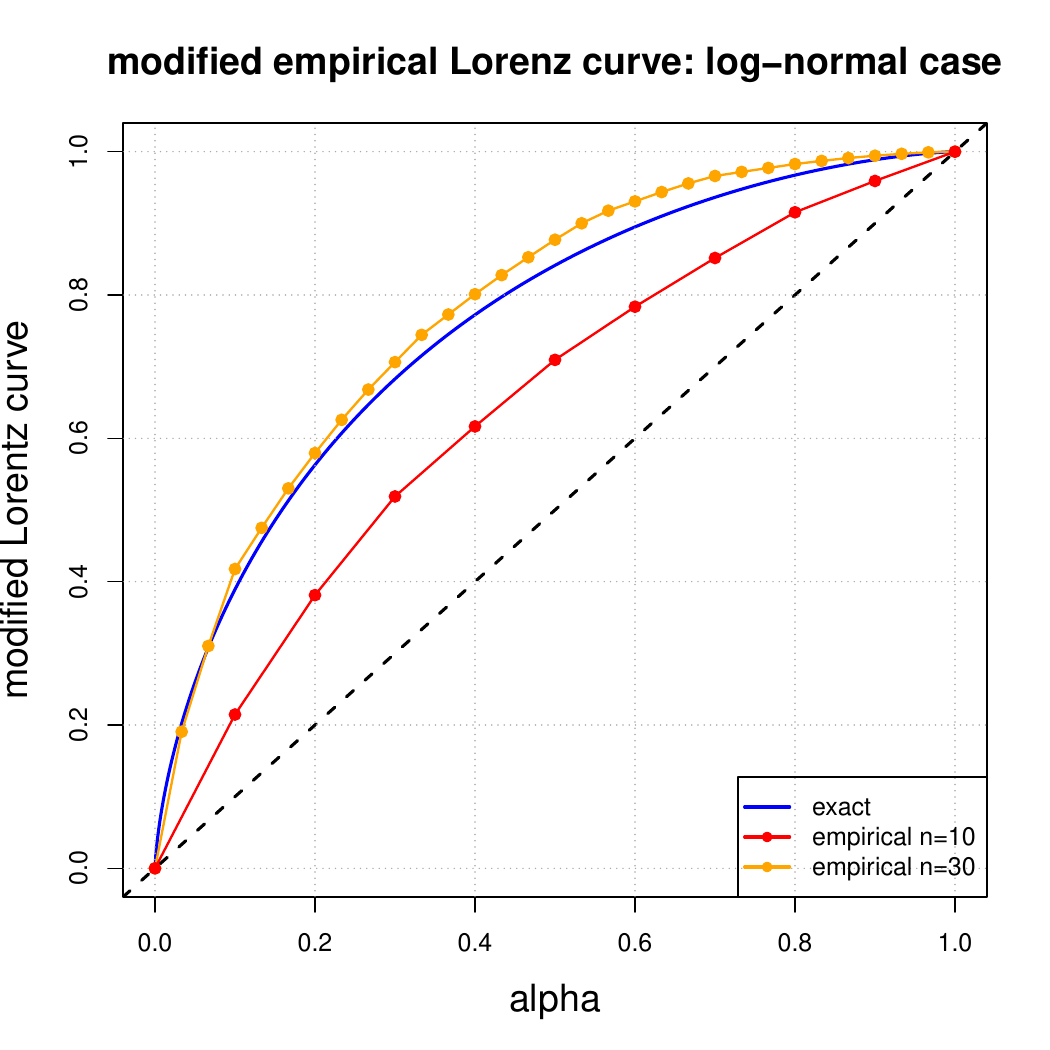}
\end{center}
\end{minipage}
\begin{minipage}[t]{0.45\textwidth}
\begin{center}
\includegraphics[width=\textwidth]{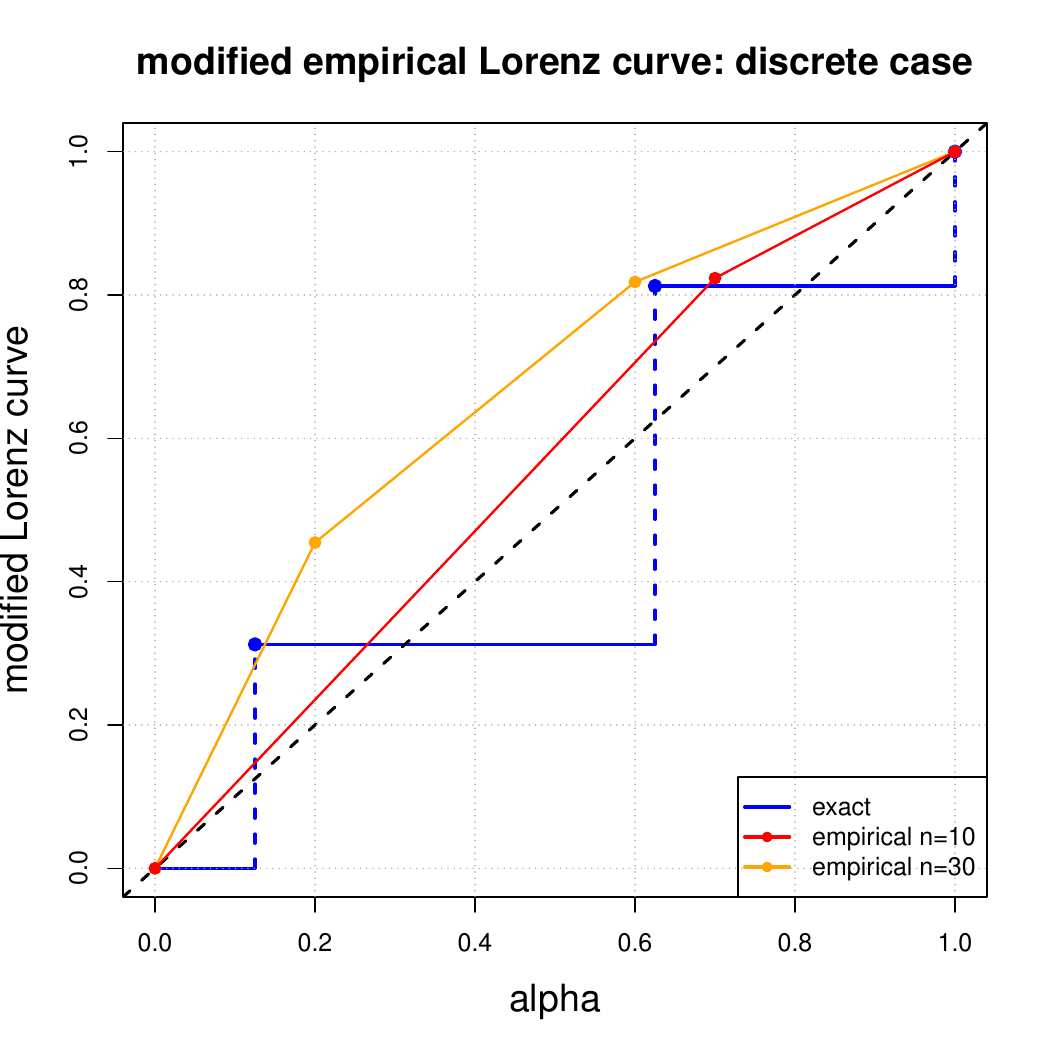}
\end{center}
\end{minipage}
\end{center}
\caption{Modified (linearly interpolated) empirical Lorenz curve $\widehat{L}^+_n$ for sample sizes $n=10, 30$: (lhs) log-normal case; (rhs) discrete case.}
\label{empirical Lorenz 2}
\end{figure}

From Figure \ref{empirical Lorenz}, one observes that the (empirical) Lorenz curves can be both above and below the diagonal dotted line. For the Gini score introduced in Section \ref{Cumulative accuracy profile}, below, it is beneficial to have a curve that does not fall below the diagonal dotted line, because one wants to compute the area enclosed by that curve and the diagonal dotted line. In the non-continuous case, this can be achieved by a linear interpolation between the discrete points (instead of a step function interpolation). Assume we have an i.i.d.~sample $(Y_i)_{i=1}^n$ from a distribution $F_Y$; this sample may have ties. Assume that this sample takes $K$ unique values $(Y^\star_k)_{k=1}^K$. This allows us to define the aggregated case weights in these unique values
\begin{equation*}
w_k^\star = \frac{1}{n}\,\sum_{i=1}^n  \mathds{1}_{\{Y_i = Y^\star_k\}},
\qquad \text{ for $1\le k \le K$.}
\end{equation*}
These aggregated case weights precisely describe the step heights in the empirical distribution of $(Y_i)_{i=1}^n$.
That is, we have the two equivalent versions of the empirical distribution
\begin{equation*}
\widehat{F}_n(y) = \frac{1}{n}\, \sum_{i=1}^n \mathds{1}_{\{ Y_i \le y \}}
=\sum_{k=1}^K w_k^\star\, \mathds{1}_{\{ Y^\star_{k} \le y \}}.
\end{equation*}
Consider the strict order statistics $Y^\star_{(1)} > \ldots >Y^\star_{(K)}$, and map the identical ordering to the case weights $(w^\star_{[k]})_{k=1}^K$ -- using the square bracket notation to indicate that this is implied by the order statistics of the responses.  The steps on the $x$-axis in the empirical graphs in Figure \ref{empirical Lorenz} are precisely in the points
\begin{equation*}
\alpha^\star_k = \sum_{j=1}^k w^\star_{[j]}, \qquad \text{ for $1 \le k \le  K$,}
\end{equation*}
and we initialize $\alpha^\star_0=0$.
The incremental step heights of the empirical Lorenz curve (on the $y$-axis) in these steps are given by
\begin{equation*}
\frac{1}{\sum_{j=1}^K w^\star_{[j]} Y^\star_{(j)}} \, w^\star_{[k]} Y^\star_{(k)},
\end{equation*}
and in the aggregated version they are equal to for $0\le k \le K$
\begin{equation*}
\widehat{L}_n(\alpha^\star_k) =\frac{1}{\sum_{j=1}^K w^\star_{[j]} Y^\star_{(j)}} \, 
\sum_{j=1}^k  w^\star_{[j]} Y^\star_{(j)}.
\end{equation*}
The orange and red dots in Figure \ref{empirical Lorenz} illustrate this set, called {\it corner set},
\begin{equation}
\label{discrete Lorenz dots}
{\cal B}^\star=\left(\alpha^\star_k, \,\widehat{L}_n(\alpha^\star_k)\right)_{k=0}^K.
\end{equation}
We now linearly interpolate between these points of the corner set ${\cal B}^\star$, that is, for
$\alpha \in (\alpha^\star_{k-1}, \alpha^\star_k]$, $1\le k \le K$, we set 
\begin{equation}\label{modified Lorenz}
\widehat{L}^+_n(\alpha)
= \widehat{L}_n(\alpha^\star_{k-1}) + \frac{\widehat{L}_n(\alpha^\star_k)-\widehat{L}_n(\alpha^\star_{k-1})}{\alpha^\star_k-\alpha^\star_{k-1}}\left(\alpha - \alpha^\star_{k-1}\right).
\end{equation}
We call this linearly interpolated version the {\it modified (empirical) Lorenz curve}. Naturally, this concept can be applied to any discrete distribution, not only to the empirical one.
The results of this linear interpolation in the empirical cases are shown by the red and orange graphs in Figure \ref{empirical Lorenz 2} (the true discrete distribution in blue is not linearly interpolated in this graph).

\medskip

\begin{figure}[htb!]
\begin{center}
\begin{minipage}[t]{0.45\textwidth}
\begin{center}
\includegraphics[width=\textwidth]{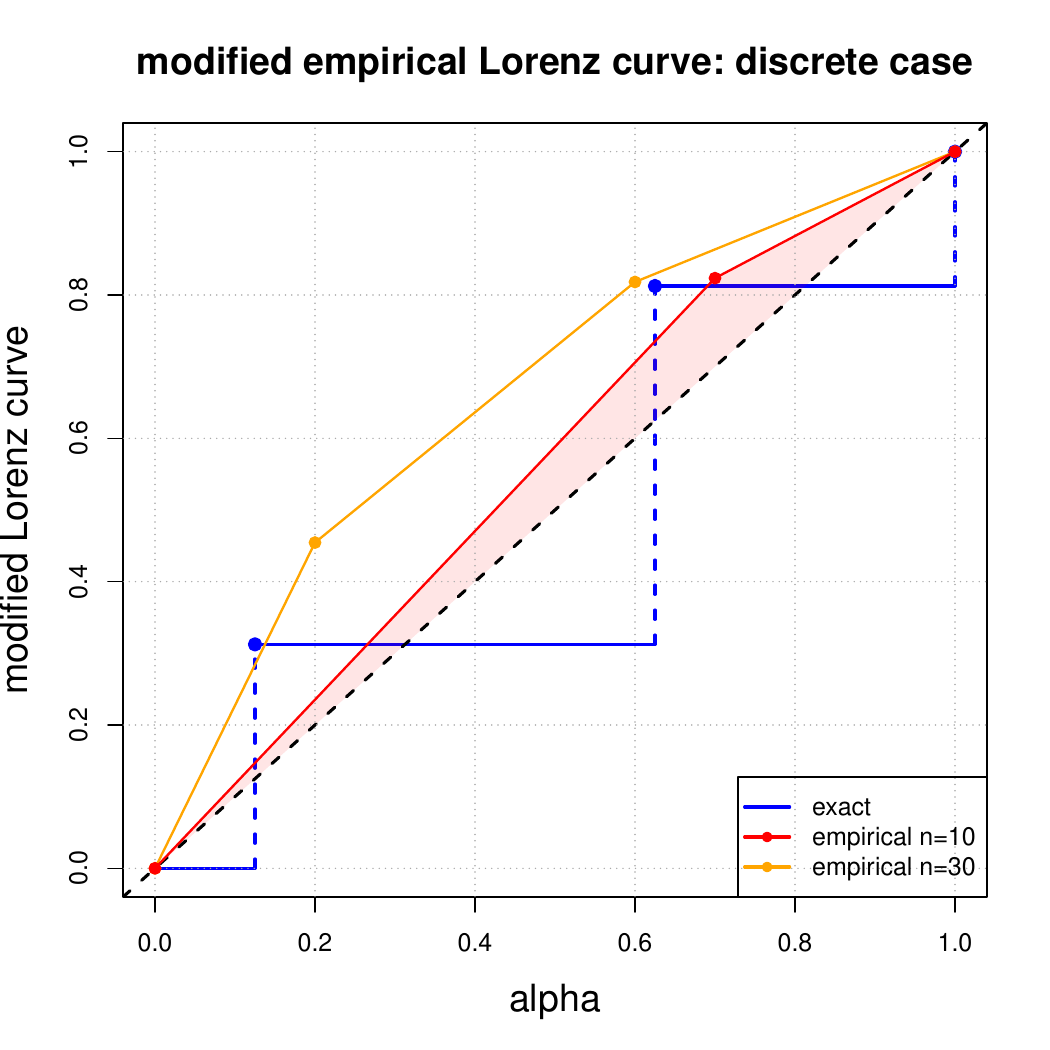}
\end{center}
\end{minipage}
\begin{minipage}[t]{0.45\textwidth}
\begin{center}
\includegraphics[width=\textwidth]{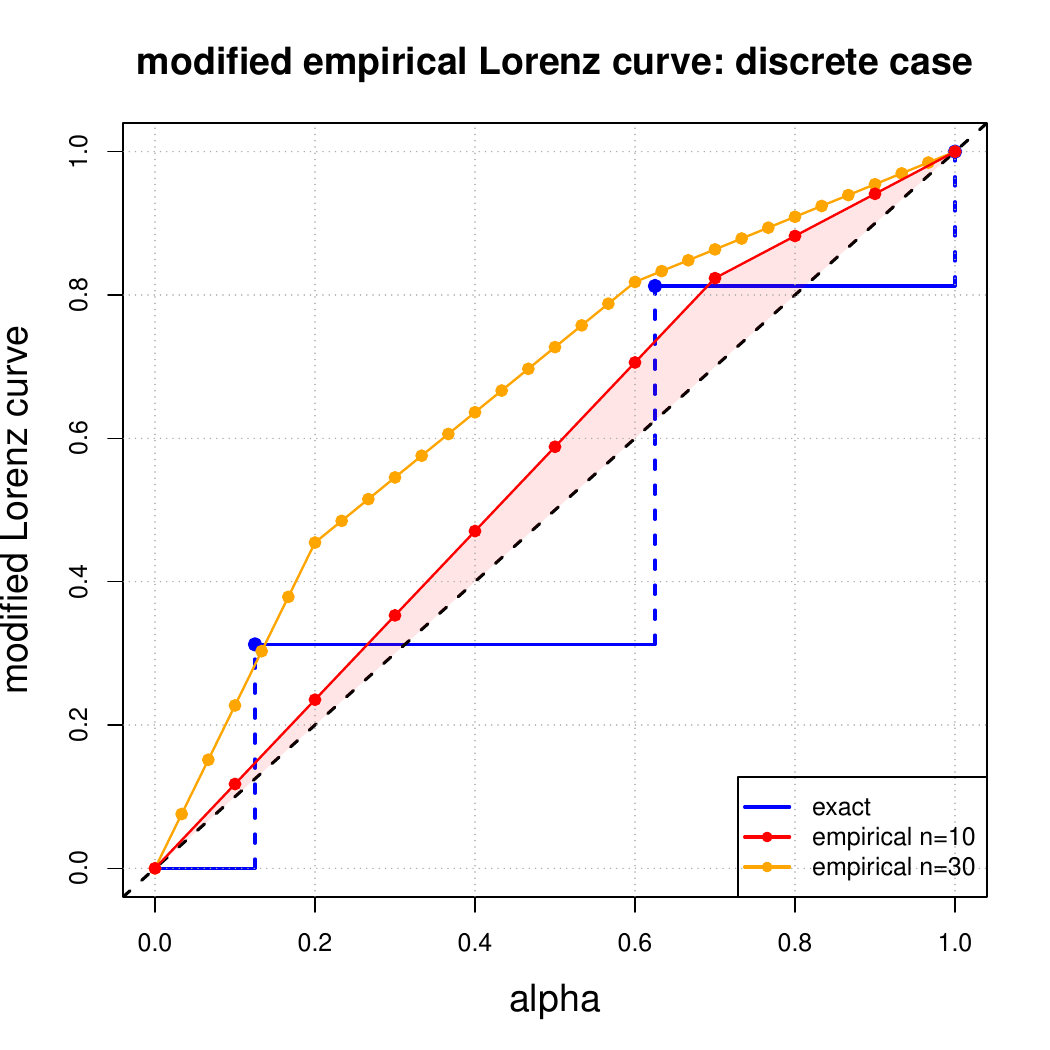}
\end{center}
\end{minipage}
\end{center}
\caption{Modified empirical Lorenz curve $\widehat{L}^+_n$ in the discrete case: (lhs) constructed on the aggregated order statistics $(Y^\star_{(k)})_{k=1}^K$ with corner set ${\cal B}^\star$, and (rhs) on the non-aggregated order statistics $(Y_{(i)})_{i=1}^n$ with corner set ${\cal B}$; the two red areas are identical.}
\label{empirical Lorenz 3}
\end{figure}

This linear interpolation comes with several advantages:
\begin{enumerate}
\item The aggregation over the unique values in the sample $(Y_i)_{i=1}^n$ is not necessary,  i.e., in case of ties they enter automatically correctly into the modified Lorenz curve. Namely, we can build the order statistics $Y_{(1)}\ge \ldots \ge Y_{(n)}$, with a deterministic rule in the ties. This allows us to consider the modified empirical Lorenz curve in the values $\alpha_i=i/n$ for $i\in \{0,\ldots, n\}$
\begin{equation}\label{corners Lorenz}
\widehat{L}^+_n(\alpha_i) =\frac{1}{\frac{1}{n}\sum_{j=1}^n Y_j} \, 
\frac{1}{n}\sum_{j=1}^i  Y_{(j)},
\end{equation}
the empty sum is set equal to zero. This gives us the corner set
\begin{equation}
\label{corner set B}
{\cal B}=\left(\alpha_i=\frac{i}{n}, \,\widehat{L}^+_n(i/n)\right)_{i=0}^n.
\end{equation}
Linear interpolation over this corner set ${\cal B}$ gives the identical curve as the modified empirical Lorenz curve \eqref{modified Lorenz}.
Figure \ref{empirical Lorenz 3} verifies that both versions -- the tie-aggregated one and the tie-non-aggregated one -- give the {\it same} piecewise linear graph.
Thus, aggregation in the ties is {\it not} necessary if we interpolate linearly over the corner set ${\cal B}$.
\item The dotted diagonal now receives an explicit meaning, namely, if all observations are identical, $Y_1=\ldots = Y_n$, the diagonal gives the modified empirical Lorenz curve, and whenever there are at least two different observations $Y_j\neq Y_k$ for some $j\neq k$, the modified empirical Lorenz curve $\widehat{L}^+_n$ lies above the dotted diagonal.
\item The area between the modified empirical Lorenz curve $\widehat{L}^+_n$ and the diagonal is always a convex set, and its area is strictly positive whenever there are at least two different observations in the sample $(Y_i)_{i=1}^n$. This is implied by the fact that the corner sets are constructed from decreasing order statistics, and it is verified in Figure \ref{empirical Lorenz 3} by the red area illustrating the smaller sample with $n=10$.
\end{enumerate}

\begin{figure}[htb!]
\begin{center}
\begin{minipage}[t]{0.45\textwidth}
\begin{center}
\includegraphics[width=\textwidth]{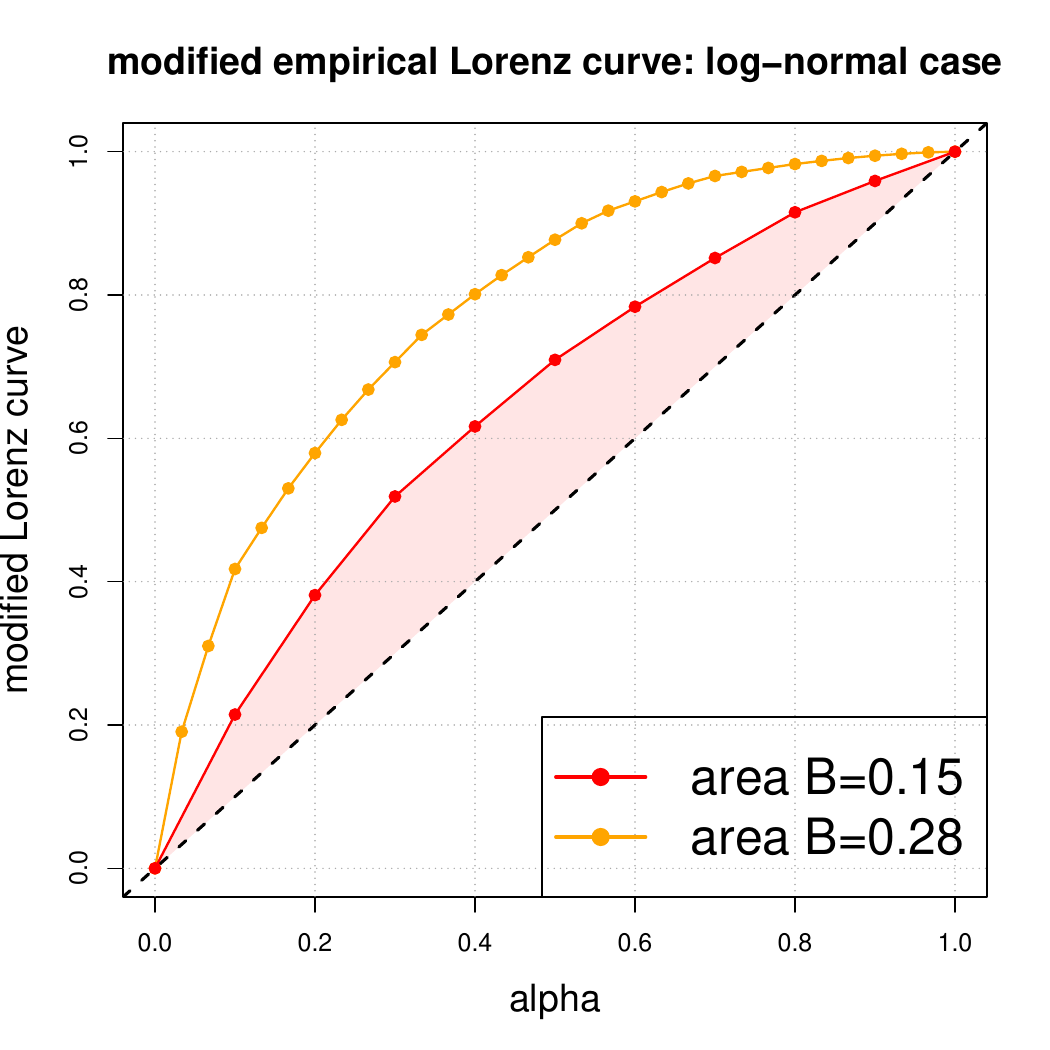}
\end{center}
\end{minipage}
\begin{minipage}[t]{0.45\textwidth}
\begin{center}
\includegraphics[width=\textwidth]{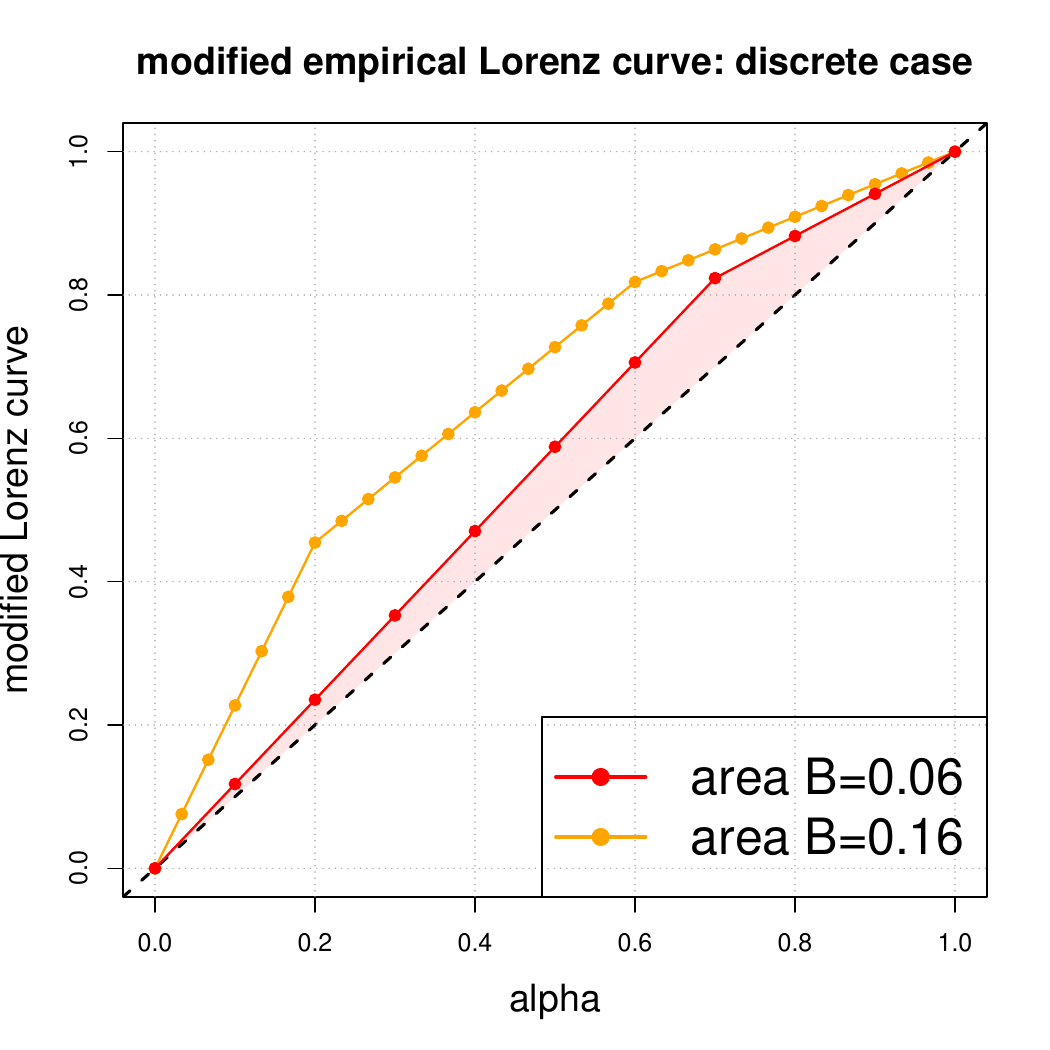}
\end{center}
\end{minipage}
\end{center}
\caption{Modified empirical Lorenz curve $\widehat{L}^+_n$ 
showing the area $B$ that is enclosed in the convex set between the diagonal dotted line and the modified empirical Lorenz curve $\widehat{L}^+_n$ for sample sizes $n=10,30$:
(lhs) log-normal case, (rhs) discrete case.}
\label{empirical Lorenz 4}
\end{figure}

In view of the last item of the above list of advantages of the modified empirical Lorenz curve, we can also easily compute the resulting convex area implied by the corner set ${\cal B}$ (above the diagonal).
In view of \eqref{corners Lorenz}-\eqref{corner set B}, we compute the size of the enclosed convex area by
\begin{equation}\label{nominator Gini}
B :=  \sum_{i=1}^n \frac{\widehat{L}^+_n(\alpha_i) + \widehat{L}^+_n(\alpha_{i-1})}{2}\,
\left(\alpha_i - \alpha_{i-1}\right) - \frac{1}{2}.
\end{equation}
The size of this area is indicated in the legend in the graphs of Figure \ref{empirical Lorenz 4}, and illustrated by the red area in the case of the smaller samples $n=10$.

\section{Cumulative accuracy profile}
\label{Cumulative accuracy profile}

\subsection{The continuous distribution case}
\label{The continuous distribution case}
We start from a random triplet $(Y,\mu,\widehat{\mu})$ characterizing an insurance policyholder. This random triplet has the following interpretation. First, we randomly select an insurance policyholder from the insurance portfolio, this is described by simulating their average claim $\mu \sim \p$ from the portfolio (population) distribution $\p$. Second, we simulate the claim $Y$, given $\mu$,  satisfying the calibration property
\begin{equation}
\label{calibration property}
\mu=\E \left[ \left. Y \right| \mu \right], \qquad \text{a.s.}
\end{equation}
Thus, $\mu$ reflects the conditionally expected claim of $Y$, and, thus, it provides the correct risk ranking in terms of conditionally expected claims.
Finally, we determine (simulate) the estimated claim $\widehat{\mu}$, and we assume that $Y$ and $\widehat{\mu}$ are conditionally independent, given $\mu$. This conditional independence reflects that we perform an out-of-sample model evaluation. We summarize these model assumptions.

\begin{ass}\label{model assumptions}
Assume the random triplet $(Y,\mu,\widehat{\mu})$ 
has the following structure:
\begin{itemize}
\item[{\rm (1)}] {\rm Calibration:} $\mu=\E[Y| \mu]$, a.s., with integrable mean $\E[|\mu|]<\infty$.
\item[{\rm (2)}] {\rm Out-of-sample evaluation:} $Y$ and $\widehat{\mu}$ are conditionally independent, given $\mu$.
\end{itemize}
\end{ass}

We can now simulate an i.i.d.~sample $(Y_i,\mu_i, \widehat{\mu}_i)_{i=1}^n$ under Assumptions \ref{model assumptions}. This provides us with a true risk ranking $(\mu_i)_{i=1}^n$ and an estimated risk ranking
$(\widehat{\mu}_i)_{i=1}^n$, and the main question that we would like to study with the Gini score is whether these two risk rankings coincide (are concordant).
We give some preliminary remarks.

\begin{rems}\normalfont
\label{remarks 1}
\begin{itemize}
\item Assumptions \ref{model assumptions} give the full modeling set-up. In practical applications, we will only be equipped with an observed i.i.d.~sample $(Y_i,\widehat{\mu}_i)_{i=1}^n$ -- the true means
$(\mu_i)_{i=1}^n$ typically are unknown -- and we would like to determine whether the estimates $(\widehat{\mu}_i)_{i=1}^n$ provide the correct risk ranking.
Since in general the true means are not available, we try to answer this question empirically by using the responses $(Y_i)_{i=1}^n$ that have been generated from these unknown means $(\mu_i)_{i=1}^n$.
\item {\it Calibration.} In the considerations below, $\mu$ describes the true conditional mean providing \eqref{calibration property}. For the risk ranking analysis, it is only important that it provides the correct ordering. 
That is, we may start from a risk ranking $\pi=g(\mu)$, for a strictly increasing function $g$. $\mu$ and $\pi$ provide the same risk ranking, and they generate the same $\sigma$-field, in fact, they are strictly comonotonic. As a consequence, the (isotonic) recalibration step promoted in \cite[Section 3.1]{ZiegelW} then provides the true conditional mean
\begin{equation*}
\mu=\E \left[ \left. Y \right| \pi \right], \qquad \text{a.s.}
\end{equation*}
Thus, knowing the correct risk ranking $\pi$, the true means $\mu$ can be inferred by this (isotonic) recalibration step.
\item {\it Out-of-sample evaluation.} The response $Y$ and the estimate $\widehat{\mu}$ are assumed to be {\it conditionally} independent, given $\mu$. Unconditionally, they should have a positive association, otherwise clearly, $\widehat{\mu}$ will not be suitable to describe the risk ranking.
\item In practical applications, typically, $\mu=\mu(\bX)$ will reflect the true regression function on the covariates $\bX$, and $\widehat{\mu}=\widehat{\mu}(\bX)$ its estimated counterpart which has been estimated on an independent learning sample ${\cal L}$. In this set-up, the simulation process of Assumptions \ref{model assumptions} is changed to simulating an insurance policyholder with covariates $\bX \sim \p$, this gives the true regression function $\bX \mapsto \mu(\bX)=\E[Y|\bX]$. This true regression function fulfills calibration \eqref{calibration property}. We then try to assess whether the estimated regression function $\bX \mapsto \widehat{\mu}(\bX)$ gives the correct risk ranking, and the dependence between $Y|_{\bX}$ and 
$\widehat{\mu}(\bX)$ is implied by the fact that both random variables are considered under the identical covariate $\bX$.
\end{itemize}
\end{rems}

The {\it cumulative accuracy profile} (CAP) modifies the (mirrored) Lorenz curve as follows
\begin{equation}\label{non-empirical CAP}
\alpha \in (0,1)~ \mapsto ~ C_{\mu, \widehat{\mu}}(\alpha) = \frac{1}{\E[\mu]} \,\E \left[ \mu \,\mathds{1}_{\{\widehat{\mu} > F_{\widehat{\mu}}^{-1}(1-\alpha)\}}\right],
\end{equation}
at the boundary of the unit interval, we set $C_{\mu, \widehat{\mu}}(0)=0$ and $C_{\mu, \widehat{\mu}}(1)=1$.
This is the mirrored version of the {\it concentration curve} of \cite[Definition 3.1]{DenuitSznajderTrufin}. The CAP describes an asymmetric concordance function, i.e., we aggregate the true means $\mu$ in the order of their estimated counterparts $\widehat{\mu}$. The more the ordering between $\mu$ and $\widehat{\mu}$ is aligned, the bigger the resulting CAP, and in the perfect ordering case, we precisely obtain the Lorenz curve $L_\mu(\alpha)=C_{\mu, \mu}(\alpha)$, thus, we have
\begin{equation}\label{order CAP and Lorenz}
C_{\mu, \widehat{\mu}}(\alpha) \le L_\mu(\alpha),
\qquad \text{ for all $\alpha \in [0,1]$,}
\end{equation}
with an equality if $\widehat{\mu}$ and $\mu$ are comonotonic.

\medskip

As explained in the first item of Remarks \ref{remarks 1}, in most applications the true mean $\mu$ is not available, but only observations $Y$ generated from that mean. This requires us to study a different version of the CAP.

\begin{lemma} \label{lemma L1}
Under Assumptions \ref{model assumptions} we have for all $\alpha \in [0,1]$
\begin{equation*}
C_{\mu, \widehat{\mu}}(\alpha) = C_{Y, \widehat{\mu}}(\alpha) = \frac{1}{\E[Y]} \,\E \left[ Y \,\mathds{1}_{\{\widehat{\mu} > F_{\widehat{\mu}}^{-1}(1-\alpha)\}}\right].
\end{equation*}
Assume $g$ is a strictly increasing function. Then, 
$C_{Y, g(\widehat{\mu})}(\alpha)=C_{Y, \widehat{\mu}}(\alpha)$ for all $\alpha\in [0,1]$.
\end{lemma}
{\Beweis
{\bf Proof.} The first statement is an immediate consequence of the model assumptions and the tower property of conditional expectations
by conditioning on $(\mu,\widehat{\mu})$. The second statement immediately follows from the fact that the two sets 
$\{\widehat{\mu} > F_{\widehat{\mu}}^{-1}(1-\alpha)\}=\{g(\widehat{\mu}) > F_{g(\widehat{\mu})}^{-1}(1-\alpha)\}$ are equal for any strictly increasing function $g$.
\EndProof}

\medskip

The first part of Lemma \ref{lemma L1} says that based on an observed i.i.d.~sample $(Y_i,\widehat{\mu})_{i=1}^n$, the CAP can be evaluated empirically. The second part of the lemma says that the CAP is a purely rank based measure in $\widehat{\mu}$ (that does not consider calibration).
Moreover, \eqref{order CAP and Lorenz} and
Lemma \ref{lemma L1} imply that 
\begin{equation}\label{relation Gini raw}
 \int_{0}^1 C_{Y, \widehat{\mu}}(\alpha) \,{\rm d}\alpha
  \le  \int_{0}^1 L_{Y}(\alpha)\, {\rm d}\alpha.
\end{equation}
This latter inequality is the core of the Gini score defined below.

\medskip

We now estimate the CAP from an i.i.d.~sample $(Y_i,\widehat{\mu}_i)_{i=1}^n$ under Assumptions \ref{model assumptions}, and additionally assuming that $\widehat{\mu}$ has a {\it continuous} distribution $F_{\widehat{\mu}}$. The latter implies that $(\widehat{\mu}_i)_{i=1}^n$ does not have ties, and we obtain the strict order statistics  
$\widehat{\mu}_{(1)}>\ldots >\widehat{\mu}_{(n)}$. We map the identical ordering to the responses $(Y_{[i]})_{i=1}^n$ -- using the square bracket notation -- i.e., the ordering of the responses is implied by the ranking of the estimates $(\widehat{\mu}_{(i)})_{i=1}^n$.
This motivates the empirical CAP
\begin{equation}\label{CAP empirical 1}
\alpha \in (0,1)~ \mapsto ~
\widehat{C}_n(\alpha) = \frac{1}{\frac{1}{n}\sum_{i=1}^n Y_i}\,
\frac{1}{n}\, \sum_{i=1}^{n-\lceil (1-\alpha)n\rceil} Y_{[i]},
\end{equation}
with initializations $\widehat{C}_n(0)=0$ and $\widehat{C}_n(1)=1$.
The main difference between \eqref{no ties empirical Lorenz}
and \eqref{CAP empirical 1} is that the former takes the order statistics of 
the responses $(Y_i)_{i=1}^n$ and the latter the order statistics of the estimates $(\widehat{\mu}_i)_{i=1}^n$.

This empirical CAP is a step function, and equivalently to above we can consider a piecewise linear modification. Similarly to \eqref{corners Lorenz}, define the corner set 
\begin{equation}\label{corners w/o ties}
{\cal A}= \left( \alpha_i=\frac{i}{n}, ~
\widehat{C}^+_n(i/n) =\frac{1}{\frac{1}{n}\sum_{j=1}^n Y_j} \, 
\frac{1}{n}\sum_{j=1}^i  Y_{[j]}\right)_{i=1}^n,
\end{equation}
and linearly interpolate between these corners to receive the modified empirical CAP $\alpha \mapsto \widehat{C}^+_n(\alpha)$ on $[0,1]$. In complete analogy, this allows us to compute the area
\begin{equation}\label{numerator Gini}
A :=  \sum_{i=1}^n \frac{\widehat{C}^+_n(\alpha_i) + \widehat{C}^+_n(\alpha_{i-1})}{2}\,
\left(\alpha_i - \alpha_{i-1}\right) - \frac{1}{2}.
\end{equation}
The estimated {\it Gini score} divides the area implied by the corner set ${\cal A}$, see \eqref{numerator Gini}, by the one implied by the corner set ${\cal B}$, see \eqref{nominator Gini}, resulting in
\begin{equation}\label{Gini score}
\operatorname{Gini}(Y_i,\widehat{\mu}_i)_{i=1}^n ~:= ~ \frac{A}{B} ~\le ~1.
\end{equation}
This assumes that not all observations $(Y_i)_{i=1}^n$ are identical, which implies $B>0$, and the upper bound is given by the fact that the Lorenz curve considers the perfect ordering which is achieved if the ranks of the responses $(Y_i)_{i=1}^n$ and the estimates $(\widehat{\mu}_i)_{i=1}^n$ are aligned.

\medskip

$\Longrightarrow$ 
This solves the case of the estimates $(\widehat{\mu}_i)_{i=1}^n$ not having any ties.

\medskip

\begin{rem}\normalfont
The nominator $B$ of the Gini score is strictly positive as soon as there are two different values in the observations. The numerator $A$ of the Gini score can have any sign. Assume that the predictions $(\widehat{\mu}_i)_{i=1}^5=(1,2,3,4,5)$ are strictly increasing and the corresponding responses are strictly decreasing $(Y_i)_{i=1}^5=(5,4,3,2,1)$; this reflects a counter-monotonic setting. In this case, we obtain a modified empirical CAP that lies below the diagonal, henceforth, $A<0$.
\end{rem}

\subsection{The non-continuous distribution case}

If the distribution $F_{\widehat{\mu}}$ is {\it not continuous}, then
the estimates $(\widehat{\mu}_i)_{i=1}^n$ likely have ties, and we do not receive a unique ordering.

\begin{equation*}
\text{
{\bf Q:}
What should be done in case of ties in $(\widehat{\mu}_i)_{i=1}^n$?}
\end{equation*}

\paragraph{First answer.} By taking inspiration of the non-empirical version \eqref{non-empirical CAP}, we simply aggregate in the ties.
Assume we have $K$ unique values $(\widehat{\mu}^\star_k)_{k=1}^K$
in $(\widehat{\mu}_i)_{i=1}^n$. This motivates the aggregate case weights
\begin{equation*}
w_k^\star = \frac{1}{n}\,\sum_{i=1}^n  \mathds{1}_{\{\widehat{\mu}_i = \widehat{\mu}^\star_k\}},
\qquad \text{ for $1\le k \le K$,}
\end{equation*}
and the average observation (empirical mean)
\begin{equation}\label{aggregation of responses}
Y_k^\star = \frac{1}{w_k^\star}\,\sum_{i=1}^n \frac{Y_i}{n}\, \mathds{1}_{\{\widehat{\mu}_i = \widehat{\mu}^\star_k\}},
\qquad \text{ for $1\le k \le K$.}
\end{equation}
We can then proceed similarly to above with the new sample $(Y^\star_k,\widehat{\mu}^\star_k, w^\star_k)_{k=1}^K$ which does not have ties in 
$(\widehat{\mu}^\star_k)_{k=1}^K$, and for the case weights we refer to Section \ref{case weights section}, below.

We argue why this is not a good option to deal with ties. Typically, the goal is to assess the risk rankings of different predictive models $(\widehat{\mu}^{(1)}_i)_{i=1}^n$ and $(\widehat{\mu}^{(2)}_i)_{i=1}^n$. By performing aggregation in the ties \eqref{aggregation of responses} of the two models, they may benefit differently from a law of large numbers. We give a small examples that demonstrates this.

Assume we have the following responses
\begin{equation*}
\left(Y_1,Y_2,Y_3,Y_4, Y_5,Y_6, Y_7,Y_8\right)=(1.99,2,3,4, 5, 6, 7, 8),
\end{equation*}
this is to say that the first two responses are difficult to distinguish, and the remaining ones are fairly well ordered. Assume we have two models for the estimates providing us with
\begin{equation*}
\left(\widehat{\mu}^{(1)}_1,  \widehat{\mu}^{(1)}_2, \widehat{\mu}^{(1)}_3, \widehat{\mu}^{(1)}_4, \widehat{\mu}^{(1)}_5, \widehat{\mu}^{(1)}_6, \widehat{\mu}^{(1)}_7, \widehat{\mu}^{(1)}_8\right) = \left(2.01,2,3,4, 5, 6, 7, 8 \right),
\end{equation*}
and 
\begin{equation*}
\left(\widehat{\mu}^{(2)}_1,  \widehat{\mu}^{(2)}_2, \widehat{\mu}^{(2)}_3, \widehat{\mu}^{(2)}_4, \widehat{\mu}^{(2)}_5, \widehat{\mu}^{(2)}_6, \widehat{\mu}^{(2)}_7, \widehat{\mu}^{(2)}_8\right) = \left(3,3,3,3,7,7,7,7 \right).
\end{equation*}
The first model $\widehat{\mu}^{(1)}$ gets the perfect ranking except on the first two responses that are difficult to distinguish, in fact, it slightly overestimates the smallest observed response. The second model $\widehat{\mu}^{(2)}$ can only rank better (3) from worse (7). Probably everyone agrees that the first model does a better job in this risk ranking.

Using averaging \eqref{aggregation of responses} for the second model with ties, we receive
\begin{equation}\label{LNN effect}
\left(Y^\star_1,Y^\star_2\right)=(2.7475, 6.5) \qquad \text{ and } \qquad 
\left((\widehat{\mu}^{(2)})^\star_1,  (\widehat{\mu}^{(2)})^\star_2\right) = \left(3,7 \right).
\end{equation}
Thus, on the aggregated scale, the second model provides the perfect ranking. Naturally, we cannot compare this aggregated statistics of the second model with the non-aggregated one of the first model, because the coarser model will typically benefit more from averaging \eqref{aggregation of responses} -- a law of large number.

\medskip

\underline{We conclude:}\\
Do not aggregate in the ties of $(\widehat{\mu}_i)_{i=1}^n$, but always consider the models on the identical (observation) scale to have a fair comparison.

\medskip

Remark that in many applied situations the estimated regression models do not have the same granularity, e.g., if one constructs two different gradient boosting machine estimators, say, only differing in their choices of the tree size or their shrinkage factors, they will have different granularity.

\paragraph{Second answer.} We do not aggregate in the ties of 
$(\widehat{\mu}_i)_{i=1}^n$, but we remain on the original observation scale providing sample size $n$. Consider the order statistics 
$\widehat{\mu}_{(1)}\ge \ldots \ge \widehat{\mu}_{(n)}$, which in case of ties does not give a unique ranking. We map this ranking to the responses $(Y_i)_{i=1}^n$, and in the ties of $(\widehat{\mu}_{i})_{i=1}^n$ we take two extreme positions. Namely, let $(Y_{[i\downarrow]})_{i=1}^n$ be the ordered sample that uses the same ordering as the order statistics of $(\widehat{\mu}_{i})_{i=1}^n$ and in the ties of $(\widehat{\mu}_{i})_{i=1}^n$ we order the responses in a decreasing order of these responses (that is, we use the responses for a suborder in the ties). For the second extreme case, we let $(Y_{[i\uparrow]})_{i=1}^n$ have an increasing suborder in the ties of $(\widehat{\mu}_{i})_{i=1}^n$ w.r.t.~the responses. This equips us with two ordered samples $(Y_{[i\downarrow]})_{i=1}^n$ and $(Y_{[i\uparrow]})_{i=1}^n$ that are ordered w.r.t.~the order statistics of $(\widehat{\mu}_{i})_{i=1}^n$ and in its ties we use the responses $(Y_i)_{i=1}^n$ for a suborder (requiring a deterministic rule of also the responses have ties in the ties of the estimates).

Completely analogously to \eqref{corners w/o ties}, we define the two corner sets ${\cal A}^\downarrow$ and ${\cal A}^\uparrow$ by setting  $\alpha_i=i/n$, for $i\in \{0,\ldots, n\}$, and
\begin{equation*}
\widehat{C}^\downarrow_n(\alpha_i) =\frac{1}{\frac{1}{n}\sum_{j=1}^n Y_j} \, 
\frac{1}{n}\sum_{j=1}^i  Y_{[j\downarrow]}
\qquad \text{ and }\qquad
\widehat{C}^\uparrow_n(\alpha_i) =\frac{1}{\frac{1}{n}\sum_{j=1}^n Y_j} \, 
\frac{1}{n}\sum_{j=1}^i  Y_{[j\uparrow]}.
\end{equation*}
An easy consequence of the selected suborders in the ties is that 
\begin{equation*}
\widehat{C}^\downarrow_n(\alpha_i) \ge \widehat{C}^\uparrow_n(\alpha_i),
\qquad \text{ for all $i\in \{0,\ldots, n\}$,}
\end{equation*}
 and naturally any other suborder in the ties is between these two bounds.
$\widehat{C}^\downarrow_n$ can be seen as a best case CAP, because the ordering within the ties is optimal w.r.t.~the responses, 
and $\widehat{C}^\uparrow_n$ then is the worst case CAP (with the least optimal ordering in the ties); in a binary context this corresponds to
\cite[Figure 6]{Fawcett}.

Because we want one single Gini score at the end, we define the mid-solution in 
$\alpha_i=i/n$ for $i\in \{0,\ldots, n\}$ by
\begin{equation} \label{mid CAP}
\widehat{C}^+_n(\alpha_i) =
\frac{1}{2}\left(\widehat{C}^\downarrow_n(\alpha_i)+  \widehat{C}^\uparrow_n(\alpha_i)\right)
=
\frac{1}{\frac{1}{n}\sum_{j=1}^n Y_j} \, 
\frac{1}{n}\sum_{j=1}^i  \frac{1}{2}\left(Y_{[j\downarrow]}+  Y_{[j\uparrow]}\right),
\end{equation}
and it is linearly interpolated between these points.

\begin{rems}\normalfont
\label{remark 3.5}
\begin{itemize}
\item
The mid-solution presented in \cite[Figure 6]{Fawcett} in the binary context differs from our proposal. This reference does not consider a suborder in the ties implied by the responses, but it rather connects the corner set obtained from the unique values $(\widehat{\mu}^\star_k)_{k=1}^K$ by straight lines. This can be interpreted as an expected sorting, because any observation within a tie occurs with the same probability, and averaging over all permutations gives a straight line instead of suborders in ties. Our mid-solution \eqref{mid CAP} generally does not give a straight line in the ties. However, we prove in Lemma \ref{lemma 1} that both solutions, the straight line of \cite{Fawcett} and our mid-solution
\eqref{mid CAP} lead to the identical value \eqref{Gini score 2} of the Gini score, i.e., the geometric objects are different but they include the same area.
\item
We prefer our solution because it is computationally more appealing: it does not require to explicitly identify the ties, but it can be implemented by first ordering for $(Y_i)_{i=1}^n$ and then for $(\widehat{\mu}_i)_{i=1}^n$, the second sorting typically keeps the order of the first one in the ties. This then gives the correct order for aggregating to receive the CAP; a code example is given in Listing \ref{RCode} in the appendix.
\item Our solution with suborders in the ties as well as the straight line solution of \cite{Fawcett} have the advantage over a random ordering in the ties, that they provide a unique (and replicable) value.
  For this reason, we do not recommend a random ordering in the ties.
\item In relation to the last item, in the case of a constant prediction $\widehat{\mu}_i\equiv \widehat{\mu}$, there is one big tie. In this case, our mid-solution provides the diagonal line, which will result in a Gini score of zero in \eqref{Gini score 2}, below.
\item The best case CAP $\widehat{C}^\downarrow_n$ and the worst case CAP $\widehat{C}^\uparrow_n$ are identical in the case where all observations in the ties are identical. This is particularly the case for the Lorenz curve that enters the nominator of the Gini score, see \eqref{Gini score 2}, below. For this reason, in case of the Lorenz curve, we do not need to consider best and worst cases, but we can directly interpolate by straight lines, see Figure \ref{empirical Lorenz 3}.
\end{itemize}
\end{rems}

This mid-solution is identical to \eqref{corners w/o ties} in the case without ties, and it extends to the case with ties in the estimates $(\widehat{\mu}_i)_{i=1}^n$.
In complete analogy to \eqref{numerator Gini}, we define in this more general case the area induced by these corner points
\begin{equation}\label{numerator Gini 2}
A =  \sum_{i=1}^n \frac{\widehat{C}^+_n(\alpha_i) + \widehat{C}^+_n(\alpha_{i-1})}{2}\,
\left(\alpha_i - \alpha_{i-1}\right) - \frac{1}{2}.
\end{equation}
The estimated Gini score is defined by
\begin{equation}\label{Gini score 2}
\operatorname{Gini}(Y_i,\widehat{\mu}_i)_{i=1}^n ~=~ \frac{A}{B} ~\le~ 1.
\end{equation}
For two models, $\widehat{\mu}^{(1)}$ and $\widehat{\mu}^{(2)}$, the second one provides the more accurate risk ranking measured by the Gini score if 
\begin{equation*}
\operatorname{Gini}(Y_i,\widehat{\mu}^{(1)}_i)_{i=1}^n
~\le ~ \operatorname{Gini}(Y_i,\widehat{\mu}^{(2)}_i)_{i=1}^n ~\le ~ 1.
\end{equation*}

\begin{lemma}
\label{lemma 1}
Consider the aggregated sample $(Y_k^\star, \widehat{\mu}_k^\star, w_k^\star)_{k=1}^K$ such that $(\widehat{\mu}_k^\star)_{k=1}^K$ does not have ties.
In complete analogy to \eqref{A1}-\eqref{A2}, below, this leads to a corner set ${\cal A}$ without ties. By connecting this corner set by straight lines we can compute the area enclosed with the diagonal line resulting precisely in the value $A$ given in \eqref{numerator Gini 2}.
\end{lemma}
This lemma states that our mid-solution \eqref{numerator Gini 2} and the straight line solution in the binary case of \cite{Fawcett} give the same Gini score, see also Remarks \ref{remark 3.5}.

\medskip

{\Beweis {\bf Proof of Lemma \ref{lemma 1}.} 
Consider the suborders $(Y_{[i\downarrow]})_{i=1}^n$ and 
$(Y_{[i\uparrow]})_{i=1}^n$ in the ties. These two suborders in the ties precisely reflect the mirrored Lorenz curve \eqref{Lorenz A2} and the original Lorenz curve \eqref{Lorenz A1}, respectively. As a consequence, the area enclosed of these two curves with the diagonal line is the same, but with opposite signs. Therefore, if we average over these two areas, we precisely obtain the same area as the one computed from the diagonal straight line. This proves the claim.
\EndProof}

\begin{example}[Gini score under ties]
\normalfont
We consider two examples with ties. Assume that the true mean $\mu$ has a uniform distribution over the set
\begin{equation*}
\left\{1,2,3,8,9,10 \right\},
\end{equation*}
and we simulate the response by a conditional Gaussian distribution $Y|\mu \sim {\cal N}(\mu,1)$. We assume to have two different estimates (models) $\widehat{\mu}^{(1)}$ and $\widehat{\mu}^{(2)}$ being deterministic functions of $\mu$, described by the following ordered samples
\begin{eqnarray*}
\mu = 1,2,3,8,9,10 & \Longrightarrow & \widehat{\mu}^{(1)}=1,2,3,10,7,8,\\
\mu = 1,2,3,8,9,10 & \Longrightarrow & \widehat{\mu}^{(2)}=2,2,2,10,7,8.
\end{eqnarray*}
That is, the first model $\widehat{\mu}^{(1)}$ has the correct risk ranking in the lowest three values, but it wrongly allocates the largest risk 10 to position 4. The second model $\widehat{\mu}^{(2)}$ has the same wrong ordering of  the biggest 3 risks, and it cannot distinguish the smallest three risks. From this we conclude that the first model $\widehat{\mu}^{(1)}$ should be preferred over the second one $\widehat{\mu}^{(2)}$ w.r.t.~the risk ranking.

The previous conclusion uses the knowledge of the true mean $\mu$, which typically is not available, and we try to find the same conclusion from a validation sample. Therefore, we generate an i.i.d.~sample $(Y_i,\mu_i,\widehat{\mu}_i)_{i=1}^n$ from this model of sample size $n=20$.

\begin{figure}[htb!]
\begin{center}
\begin{minipage}[t]{0.45\textwidth}
\begin{center}
\includegraphics[width=\textwidth]{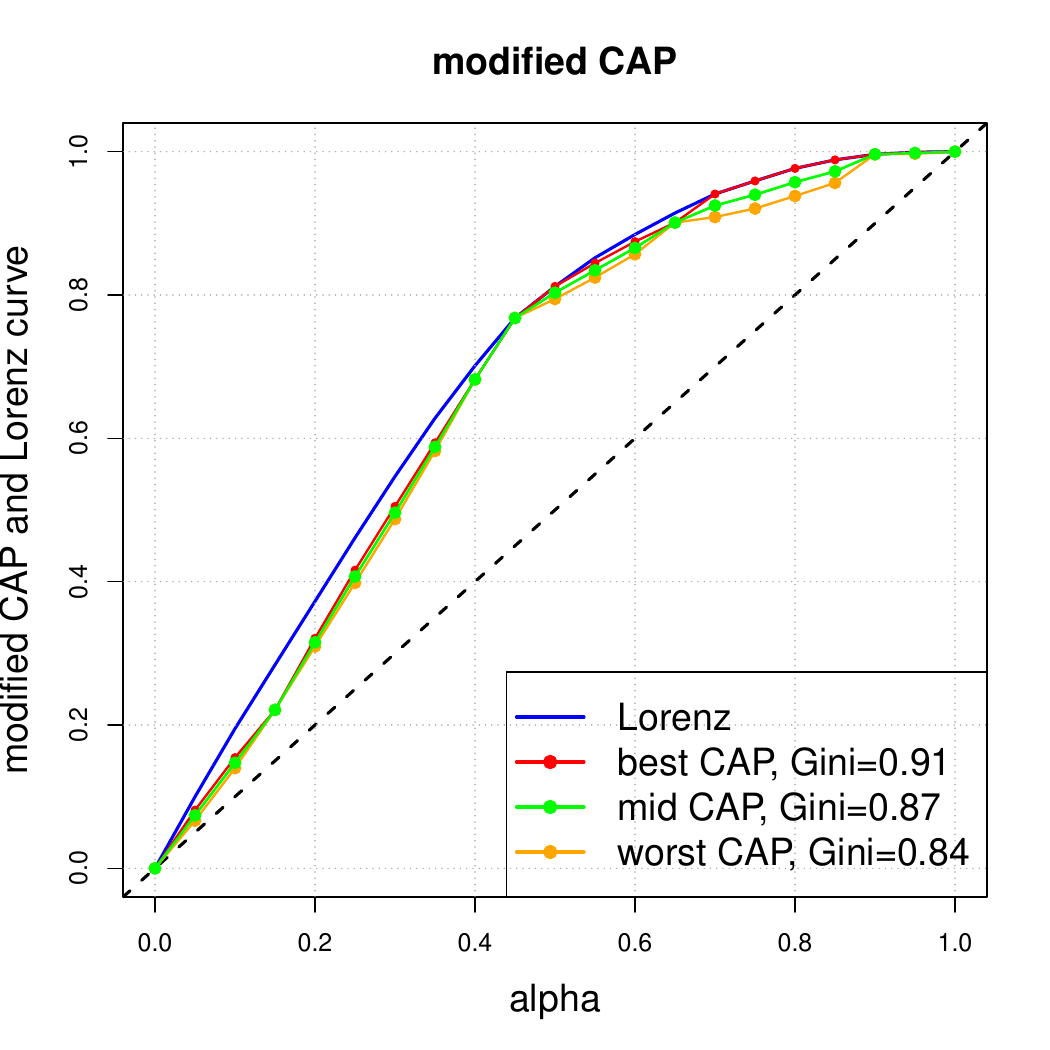}
\end{center}
\end{minipage}
\begin{minipage}[t]{0.45\textwidth}
\begin{center}
\includegraphics[width=\textwidth]{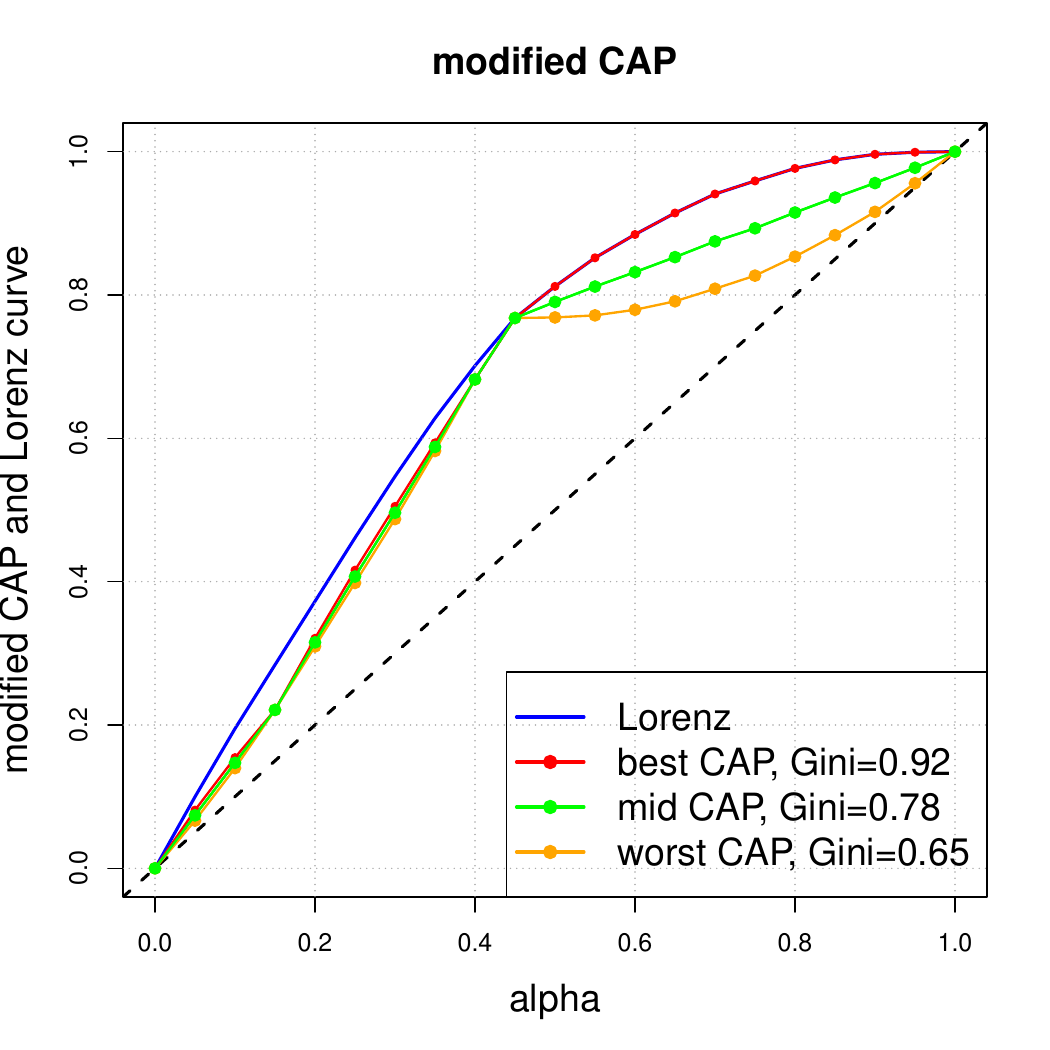}
\end{center}
\end{minipage}
\end{center}
\caption{Gini score evaluated on a sample size of $n=20$:
(lhs) first model $\widehat{\mu}^{(1)}$, (rhs) second model $\widehat{\mu}^{(2)}$.}
\label{fig Gini 1}
\end{figure}

The results of the two estimation models $\widehat{\mu}^{(1)}$ and $\widehat{\mu}^{(2)}$ are shown in Figure \ref{fig Gini 1} (lhs) and (rhs), respectively.
The blue curve shows the modified (linearly interpolated) Lorenz curve
\eqref{corners Lorenz}-\eqref{corner set B}, which determines the nominator $B$ of the Gini score, see \eqref{nominator Gini}. The red and the orange curves are best CAP $\widehat{C}^\downarrow_n$ and the worst CAP $\widehat{C}^\uparrow_n$, respectively, and the green line
gives the mid-solution CAP $\widehat{C}^+_n$, see \eqref{mid CAP},  that is used to compute the numerator A of the Gini index, see \eqref{numerator Gini 2}-\eqref{Gini score 2}. This results in Gini scores
\begin{equation*}
\operatorname{Gini}(Y_i,\widehat{\mu}^{(1)}_i)_{i=1}^n =0.87
~> ~ \operatorname{Gini}(Y_i,\widehat{\mu}^{(2)}_i)_{i=1}^n =0.78.
\end{equation*}
Thus, we conclude from these Gini scores that the first model $\widehat{\mu}^{(1)}$ gives the more accurate risk ranking, which is the same conclusion as we have drawn above.

An important insight from Figure \ref{fig Gini 1} is also the following. We have also computed that Gini scores if we would focus on the best CAPs $\widehat{C}^\downarrow_n$ for the two models, red graphs in the two plots.
We note that this best CAP would give (a wrong) preference to the second model that cannot distinguish the lower three risk rankings correctly. The reason for this is that the equal scores give a bigger tie in the second model 
$\widehat{\mu}^{(2)}$ against the first model $\widehat{\mu}^{(1)}$, and using the optimal suborder in this tie gives an unjustified preference to the second model (Gini score 0.92 vs.~0.91 in the first model). The effect here is rather similar to law of large numbers effect \eqref{LNN effect}, and it says that for model selection, we should use the mid-solution \eqref{mid CAP}, to ensure that both models are judge in the same granularity.
\EndExample
\end{example}

The implementation of our mid-solution \eqref{numerator Gini 2}
is straightforward, code is given in the appendix; see Listing \ref{RCode}, below. Using standard software needs some care because some of these tools consider the best case CAP, which is not fully suitable as argued in the previous example.

\section{Case weights}
\label{case weights section}

Insurance data often includes so-called {\it case weights}. In fact, an insurance policyholder is characterized by a random quadruplet $(Y,\mu,\widehat{\mu},w)$, that is, the random triplet of Section
\ref{The continuous distribution case} is extended by a positive case weight $w>0$. This case weight comes from the fact that insurance liabilities and claims are often measured relative to an exposure measure, such as a time exposure, a total insurance sum, a building value, number of employees insured, etc. The typical case is the one of claims frequency, in that case, $Y$ models a claims frequency and $wY$ the total number of claims in that period of length $w$. The prediction $\widehat{\mu}$ is then on the level of a premium rate, and $w\widehat{\mu}$ gives the effectively charged premium for the claim $wY$ in the time interval of length $w$.

From a modeling perspective, this is supported by the exponential dispersion family (EDF) of distributions. This is the most popular class of distributions for regression modeling, e.g., generalized linear models (GLMs) are based on the EDF, and it contains the popular examples of the Gaussian, Poisson, negative binomial, gamma, inverse Gaussian and Tweedie's distributions. The standard representation within the EDF is precisely the one, where $Y$ is a scaled quantity with exposure weight $w>0$; this is also called the reproductive form of the random variable $Y$, see \cite{Jorgensen}.

Having an i.i.d.~sample $(Y_i,\mu_i,\widehat{\mu}_i,w_i)_{i=1}^n$, we may analyze the problem of a correct risk ranking completely analogously as in the previous Section \ref{Cumulative accuracy profile}. There is one question, though: Should we perform the risk ranking in the scaled, $Y$, or the unscaled, $wY$, version. If we perform this risk ranking in the unscaled version, we consider the case weight-adjusted i.i.d.~sample 
$(w_iY_i,w_i\mu_i,w_i\widehat{\mu}_i)_{i=1}^n$. Based on this unscaled i.i.d.~sample we proceed completely analogously to Section \ref{Cumulative accuracy profile}.

We argue why one should rather consider the scaled version
$(Y_i,\mu_i,\widehat{\mu}_i,w_i)_{i=1}^n$. Often, the case weights $w_i$ are deterministic quantities that are fixed in the insurance contracts at their inception. E.g., we may have a small business company that wants to purchase accident insurance for their $w_1=100$ workers and we may have a big company that wants to insure $w_2=1000$ workers against accidents. Under the assumption that the claims ratios $\mu_i$ do not depend on the case weights, we should charge $w_1 \mu_1$ to the first company and $w_2 \mu_2$ to the second company. Assume that the business of the second company is more prone to accidents (e.g., office work vs.~construction work), and we may assume $\mu_1 = 1$ and $\mu_2=2$, i.e., $\mu_1 < \mu_2$. Thus, $(\mu_i)_{i=1}^n$ describes the correct propensity to claims which needs to be scaled by the sizes (volumes) of the insurance contracts. Assume in the previous example that we have predictors $\widehat{\mu}_1 = 1.75$ and $\widehat{\mu}_2=1.50$, providing the wrong risk ranking $\widehat{\mu}_1>\widehat{\mu}_2$. This wrong risk ranking will be over-ruled in the unscaled case, i.e., 
$w_1\widehat{\mu}_1=175<w_2\widehat{\mu}_2=1500$, and we will not detect that our tariff scheme
$(\widehat{\mu}_i)_{i=1}^n$ does not match the correct risk ranking of the true means
$(\mu_i)_{i=1}^n$. For this reason, we prefer the scaled quantities 
$(Y_i,\mu_i,\widehat{\mu}_i,w_i)_{i=1}^n$ because it brings us closer to the ground truth of the true means $(\mu_i)_{i=1}^n$.

Integrating case weights into the Gini score is done by replacing the identical empirical weights of $1$ by the corresponding case weights. In the first step, we compute the nominator of the Gini score. This is obtained by modifying the corner set ${\cal B}$ given in \eqref{corner set B} as follows. Build the order statistics $Y_{(1)} \ge \ldots \ge Y_{(n)}$ on the scaled responses, and define its implied order $(w_{[i]})_{i=1}^n$ on the case weights. On the $x$-axis of the Lorenz curve, we set for $0\le i \le n$
\begin{equation}\label{case weight scale}
\alpha_i  = \frac{1}{ \sum_{j=1}^n w_j}\, \sum_{j=1}^i w_{[j]}, 
\end{equation}
and on the $y$-axis
\begin{equation}\label{corners Lorenz weight}
\widehat{L}^+_n(\alpha_i) =\frac{1}{\sum_{j=1}^n w_jY_j} \, 
\sum_{j=1}^i  w_{[j]} Y_{(j)}.
\end{equation}
This gives us the new corner set ${\cal B}=(\alpha_i, \,\widehat{L}^+_n(\alpha_i))_{i=0}^n$ for the weighted quantities. By linear interpolation, we compute the nominator of the Gini score by
\begin{equation}\label{nominator Gini weight}
B =  \sum_{i=1}^n \frac{\widehat{L}^+_n(\alpha_i) + \widehat{L}^+_n(\alpha_{i-1})}{2}\,
\left(\alpha_i - \alpha_{i-1}\right) - \frac{1}{2}.
\end{equation}
Naturally, this includes \eqref{nominator Gini} using the specific choice $w_i\equiv 1$. 

For the numerator of the Gini score,
we consider the order statistics  $\widehat{\mu}_{(1)}\ge \ldots \ge \widehat{\mu}_{(n)}$, and in case of ties we use the best and worst suborder implied by the responses $(Y_i)_{i=1}^n$. This implies the labeling of the responses $(Y_{[i\downarrow]})_{i=1}^n$ and $(Y_{[i\uparrow]})_{i=1}^n$, and we map this ordering also to the case weights 
$(w_{[i\downarrow]})_{i=1}^n$ and $(w_{[i\uparrow]})_{i=1}^n$.
Then, we define the two corner sets ${\cal A}^\downarrow$ and ${\cal A}^\uparrow$ by setting  
\begin{equation}\label{A1}
\alpha^\downarrow_i  = \frac{1}{ \sum_{j=1}^n w_j}\,\sum_{j=1}^i w_{[j\downarrow]}
\qquad \text{ and }\qquad
\alpha^\uparrow_i  = \frac{1}{ \sum_{j=1}^n w_j}\,\sum_{j=1}^i w_{[j\uparrow]}, 
\end{equation}
for $i\in \{0,\ldots, n\}$, and
\begin{equation}\label{A2}
\widehat{C}^\downarrow_n(\alpha^\downarrow_i) =\frac{1}{\sum_{j=1}^n w_jY_j} \, 
\sum_{j=1}^i  w_{[j\downarrow]}Y_{[j\downarrow]}
\qquad \text{ and }\qquad
\widehat{C}^\uparrow_n(\alpha^\uparrow_i) =\frac{1}{\sum_{j=1}^n w_jY_j} \, 
\sum_{j=1}^i  w_{[j\uparrow]}Y_{[j\uparrow]},
\end{equation}
again we initialize these quantities to zero in zero and to one in one.
This gives us a best and worst case area, respectively,
\begin{eqnarray}\label{A3a}
A^\downarrow &=&  \sum_{i=1}^n \frac{\widehat{C}^\downarrow_n(\alpha^\downarrow_i) + \widehat{C}^\downarrow_n(\alpha^\downarrow_{i-1})}{2}\,
\left(\alpha^\downarrow_i - \alpha^\downarrow_{i-1}\right) - \frac{1}{2},
\\\label{A3b}
A^\uparrow &=&  \sum_{i=1}^n \frac{\widehat{C}^\uparrow_n(\alpha^\uparrow_i) + \widehat{C}^\uparrow_n(\alpha^\uparrow_{i-1})}{2}\,
\left(\alpha^\uparrow_i - \alpha^\uparrow_{i-1}\right) - \frac{1}{2}.
\end{eqnarray}
The estimated Gini score under case weights is defined by
\begin{equation}\label{Gini score 2 weight}
\operatorname{Gini}(Y_i,\widehat{\mu}_i, w_i)_{i=1}^n ~=~ \frac{(A^\downarrow+A^\uparrow)/2}{B} ~\le~ 1.
\end{equation}
This coincides with \eqref{Gini score 2} for identical case weights $w_i \equiv {\rm constant}>0$. Note that we cannot directly merge the best and worst case corner sets of the CAP because the aggregated weights
$(\alpha^\downarrow_i)_{i=1}^n$ and $(\alpha^\uparrow_i)_{i=1}^n$ may take different values on the $x$-axis. The code for \eqref{Gini score 2 weight} is given in Listing \ref{RCode} in the appendix.

\section{Real data example}
\label{sec: Real data example}
We give an example on the well-known French motor third-party liability (MTPL) dataset of \cite{Dutang}. We consider claims frequencies for $Y$ and we use the same data cleaning and the learning-test partition as in \cite{WM2023}.\footnote{The cleaned version can be downloaded from \url{https://aitools4actuaries.com/}.}

\begin{table}[h]
\footnotesize
\centering
\caption{Dataset characteristics.}
\label{tab:dataset}
\begin{tabular}{lcc}
\toprule
\textbf{Characteristic} & \textbf{Learning set} ${\cal L}$ & \textbf{Test set} ${\cal T}$ \\
\midrule
Number of policies & 610,206 & 67,801 \\
Total exposure (years) & 322,392 & 35,967 \\
Number of claims & 23,738 & 2,645 \\
Average frequency & 7.36\% & 7.35\% \\
Minimal number of claims per policy & 0 & 0 \\
Maximal number of claims per policy & 5 & 5 \\
\midrule
\multicolumn{3}{l}{\textbf{Covariate and response description}} \\
\midrule
\multicolumn{3}{l}{Categorical (4): Area, VehGas, VehBrand, Region} \\
\multicolumn{3}{l}{Continuous (5): VehPower, VehAge, DrivAge, BonusMalus, Density} \\
\multicolumn{3}{l}{Target: ClaimNb (claim count)} \\
\multicolumn{3}{l}{Exposure: Exposure (in yearly units)} \\
\bottomrule
\end{tabular}
\end{table}

The characteristics of the data are given in Table \ref{tab:dataset}. We fit two different GLMs on the learning dataset ${\cal L}$. The two GLMs use different sets of covariates: the first GLM, called {\tt glm1}, is a more granular regression model, and the second GLM, called {\tt glm2}, is a more crude regression model, only considering part of the covariates. Naturally, we expect that first GLM provides a better risk ranking because it considers more (granular) covariate information. We fit these two regression models on the learning sample ${\cal L}$, which equips us with the two regression functions 
\begin{equation*}
\bX \mapsto \widehat{\mu}^{\tt glm1}(\bX)
\qquad \text{ and }\qquad \bX \mapsto \widehat{\mu}^{\tt glm2}(\bX).
\end{equation*}
All the subsequent analysis is done out-of-sample on the test sample ${\cal T}$, only. In particular, we receive the two test samples
${\cal T}^{\tt glm1}=(Y_i, \widehat{\mu}_i^{\tt glm1},w_i)_{i=1}^n$ and 
${\cal T}^{\tt glm2}=(Y_i, \widehat{\mu}_i^{\tt glm2},w_i)_{i=1}^n$, where $(Y_i,w_i)$ are the out-of-sample observed frequencies and exposures of the test instances $1\le i \le n$ in ${\cal T}$. The two predictors are obtained by considering the previously fitted regression functions on the corresponding out-of-sample instances with covariates $\bX_i$, that is,
\begin{equation*}
\widehat{\mu}_i^{\tt glm1}:=\widehat{\mu}^{\tt glm1}(\bX_i)
\qquad \text{ and }\qquad
\widehat{\mu}_i^{\tt glm2}:=\widehat{\mu}^{\tt glm2}(\bX_i).
\end{equation*}
If we compute the out-of-sample Poisson deviance losses of the two models, we receive
\begin{equation*}
\text{Poisson deviance of $\widehat{\mu}^{\tt glm1}$}=0.457 
\qquad \text{ and } \qquad
\text{Poisson deviance of $\widehat{\mu}^{\tt glm2}$}= 0.475.
\end{equation*}
Thus, the first GLM clearly performs better in terms of the Poisson deviance loss; for theoretical background on loss scoring we refer to \cite[Formula (1.26) and Table 2.2]{AITools}.

\begin{figure}[htb!]
\begin{center}
\begin{minipage}[t]{0.6\textwidth}
\begin{center}
\includegraphics[width=\textwidth]{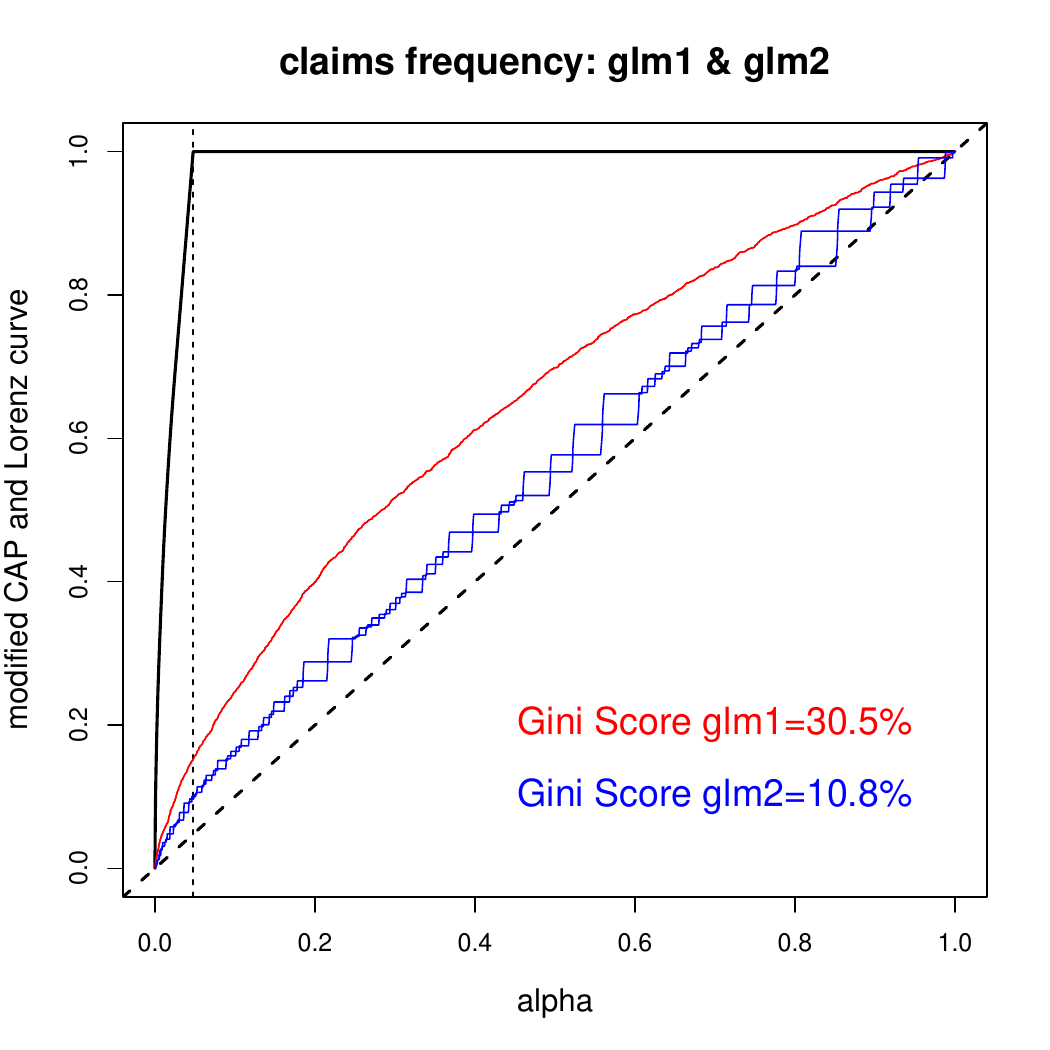}
\end{center}
\end{minipage}
\end{center}
\vspace{-.8cm}
\caption{French MTPL data: Gini scoring of the two different GLMs:
{\tt glm1} in red color and {\tt glm2} in blue color. The blue and red curves show the best and worst case CAP, the black solid line gives the Lorenz curve.}
\label{Gini MTPL}
\end{figure}

Our goal is to understand, how these two models perform in terms of the risk ranking computed by the Gini score. We start by computing the Lorenz curve
\eqref{case weight scale}-\eqref{corners Lorenz weight}
which provides the nominator $B$ of the Gini score. We note that this MTPL claims frequency data faces the typical class imbalance, meaning that insurance policies with zero claims $Y_i=0$ are much more common than policies with claims $Y_i>0$; this is also reflected in the low empirical frequency of roughly $7.35\%$, see Table \ref{tab:dataset}. This class imbalance results in a steep Lorenz curve until it hits the level one, and then it is constantly equal to one; see black solid curve in Figure \ref{Gini MTPL}. The thin black vertical dotted line is at
\begin{equation*}
\frac{1}{\sum_{i=1}^n w_i} \sum_{i=1}^n w_i  \,\mathds{1}_{\{ Y_i>0\}}=
4.77\%,
\end{equation*}
this is the relative exposure of instances having claims.

Next, we compute the best and worst case empirical CAPs \eqref{A1}-\eqref{A2} which provide the nominators $A^\downarrow$ and $A^\uparrow$ of the Gini score. A consequence of the class imbalance is that the pure randomness (irreducible risk) dominates the systematic regression structure, typically, resulting in out-of-sample losses that are dominated by irreducible risk. This also manifests in comparably small Gini scores measured  by the observation version $C_{Y,\widehat{\mu}}(\alpha)$ of the CAP, because the randomness in $Y$ dominates this quantity. We compute the worst and the best case CAPs in both GLMs, {\tt glm1} and {\tt glm2}, and they are illustrated in red and blue color in Figure \ref{Gini MTPL}. {\tt glm2} in blue color is a coarse-grained GLM having many ties in the predictions $\widehat{\mu}_i^{\tt glm2}$, this results in CAP graphs in which we can clearly distinguish the best from the worst case CAP, see Figure \ref{Gini MTPL}. For the fine-grained {\tt glm1} in red color, the differences are invisible, because we have less (big) ties and observations in the ties are more similar.

Computing the Gini score \eqref{Gini score 2 weight} from these graphs
\begin{equation*}
\operatorname{Gini}(Y_i,\widehat{\mu}^{\tt glm1}_i, w_i)_{i=1}^n=30.5\%
\qquad \text{ and } \qquad
\operatorname{Gini}(Y_i,\widehat{\mu}^{\tt glm2}_i, w_i)_{i=1}^n=10.8\%,
\end{equation*}
we give clear preference to the first GLM in terms of the risk ranking; the code is given in Listing \ref{RCode} in the appendix.

\section{Conclusion}
\label{sec: Conclusion}
The Gini score is a popular tool for model validation and model selection in statistical modeling and machine learning. It is a purely rank based score that allows one to assess risk rankings, that is, it only assesses the risk rankings of the predictions but not their levels (calibration); see \cite{WEAJ}.
Typically, this Gini score is considered under the assumption that the resulting risk ranking does not have ties. This paper discusses the treatment of ties as well as the consideration of (additional) case weights. In particular, in case of ties, we propose to consider two different suborders in the ties, and averaging over the two results gives us a unique Gini score. The technical construction of this Gini score is different from the straight line solution of \cite{Fawcett} presented in a binary context, but the resulting Gini score values are identical. Our second main statement is that if we want to compare the risk rankings of two different models, their Gini scores should be computed on the same granularity, otherwise one of the two models may benefit (more) from a law of large numbers which will result in incomparable Gini scores.

\bigskip

{\small %\baselineskip.5em
\renewcommand{\baselinestretch}{.51}
}

\newpage

\appendix

\section{Code}

\begin{lstlisting}[float=h!,frame=tb,caption={R code for Gini score computation with ties and case weights.},label=RCode]
Gini.Score <- function(pred, obs, weight) {
  
  df      <- data.frame(pred=pred,  obs=obs, weight=weight)
  N0      <- nrow(df)
  
  # 1st ordering in ties for numerator A
  df      <- df[order(df$obs, decreasing = FALSE), ]
  df      <- df[order(df$pred, decreasing = TRUE), ]
  alphaA  <- cumsum(df$weight) / sum(df$weight)
  CAPA    <- cumsum(df$obs*df$weight)/sum(df$obs*df$weight)
  A1      <- sum((CAPA + c(0,CAPA[-N0]))/2 * (alphaA - c(0,alphaA[-N0]))) - 1/2
  
  # 2nd ordering in ties for numerator A
  df      <- df[order(df$obs, decreasing = TRUE), ]
  df      <- df[order(df$pred, decreasing = TRUE), ]
  alphaA  <- cumsum(df$weight) / sum(df$weight)
  CAPA    <- cumsum(df$obs*df$weight)/sum(df$obs*df$weight)
  A2      <- sum((CAPA + c(0,CAPA[-N0]))/2 * (alphaA - c(0,alphaA[-N0]))) - 1/2
  
  # nominator B
  df      <- df[order(df$obs, decreasing = TRUE), ]
  alphaB  <- cumsum(df$weight) / sum(df$weight)
  CAPB    <- cumsum(df$obs*df$weight)/sum(df$obs*df$weight)
  B       <- sum((CAPB + c(0,CAPB[-N0]))/2 * (alphaB - c(0,alphaB[-N0]))) - 1/2
  
  (A1+A2)/(2*B)

}

\end{lstlisting}

\end{document}